\documentclass[journal]{IEEEtran}

\usepackage{cite} 

\usepackage[cmex10]{amsmath}
\usepackage{url}

\usepackage{natbib}

\usepackage{graphicx}
\usepackage{multirow}
\usepackage{verbatim}  

\def\figref#1{Fig.~\ref{#1}}
\def\eqnref#1{Eq.~\ref{#1}}
\def\secref#1{Sec.~\ref{#1}}

\usepackage{multirow}
\usepackage{verbatim}  

\usepackage{flushend}

\usepackage{subfig} 

\usepackage{wasysym}
\newcommand{\cross}{\ensuremath{\times}}

\usepackage{color}

\newcommand{\mycaption}[1]{\caption{\textit{#1}}}
\newcommand{\mysubsubsection}[1]{\subsubsection{\textbf{#1}}}
\newcommand{\mytexttt}[1]{\textit{#1}}

\usepackage{tikz}
\usepackage{schemabloc}
\usetikzlibrary{shapes,arrows}

\tikzstyle{nooutlineblock} = [rounded rectangle, minimum height=1em, minimum width=1em]
\tikzstyle{clearblock} = [draw, rounded rectangle=0.01, thick, minimum height=1em, minimum width=1em]
\tikzstyle{myline} = [-, thick, rounded corners = 5.0]
\tikzstyle{myarrow} = [-latex, thick, rounded corners=5.0]
\tikzstyle{myarrow_notrounded} = [-latex, thick]
\tikzstyle{input} = [coordinate]
\tikzstyle{output} = [coordinate]
\tikzstyle{pinstyle} = [pin edge={transparent}, pin distance=0.05cm]

\title{\LARGE Manipulation in Clutter with Whole-Arm Tactile Sensing}

\begin{document}

\author{Advait Jain$^1$ \qquad Marc D. Killpack$^1$ \qquad
    \qquad Aaron Edsinger$^2$ \qquad Charles C.
        Kemp$^1$\\
$^1$Healthcare Robotics Lab, Georgia Tech \qquad $^2$Meka Robotics%
\vspace{-0.7cm}
}

\maketitle

\begin{abstract}

We begin this paper by presenting our approach to robot manipulation,
which emphasizes the benefits of making contact with the world
across the entire manipulator.  We assume that low contact forces are
benign, and focus on the development of robots that can control their
contact forces during goal-directed motion. Inspired by biology, we
assume that the robot has low-stiffness actuation at its joints, and
tactile sensing across the entire surface of its manipulator. We then
describe a novel controller that exploits these assumptions. The
controller only requires haptic sensing and does not need an explicit
model of the environment prior to contact. It also handles multiple
contacts across the surface of the manipulator. The controller uses
model predictive control (MPC) with a time horizon of length one, and
a linear quasi-static mechanical model that it constructs at each time
step.  We show that this controller enables both real and simulated
robots to reach goal locations in high clutter with low contact
forces.  Our experiments include tests using a real robot with a novel
tactile sensor array on its forearm reaching into simulated foliage
and a cinder block. In our experiments, robots made contact across
their entire arms while pushing aside movable objects, deforming
compliant objects, and perceiving the world. 

\vspace{-0.2cm}
\end{abstract}

\IEEEpeerreviewmaketitle

\section{Introduction}


Research on robot manipulation has often emphasized collision free
motion with occasional contact restricted to the robot's end
effector. In essence, most of the manipulator's motion is intended to
be free-space motion and unintended contact is considered to be a
failure of the system. In contrast, animals often appear to treat
contact between their arms and the world as a benign and even
beneficial event that does not need to be avoided. For example, humans
make extensive contact with their forearms even during mundane tasks,
such as eating or working at a desk.

Within this paper, we present progress towards new foundational
capabilities for robot manipulation that take advantage of contact
across the entire arm. Our primary assumption is that, for a given
robot and environment, contact forces below some value have no
associated penalty. For example, when reaching into a bush, moderate
contact forces are unlikely to alter the robot's arm or the bush in
undesirable ways. Likewise, even environments with fragile objects,
such as glassware on a shelf, can permit low contact forces. While
some situations merit strict avoidance of contact with an object, we
consider these to be rare, and instead focus on default
strategies that allow contact.


In order to keep contact forces low enough to avoid penalties, we
further assume that the robot arm has compliant actuation at its
joints and tactile sensing across all of its surfaces. Low-stiffness
compliant actuation can reduce contact forces due to perturbations,
error, and other sources. Tactile sensing enables direct monitoring of
contact forces (and the distribution of contact forces). These assumed
hardware capabilities for the robot's actuation and sensing are also
analogous to capabilities found in animals.

Our main contribution in this paper is a novel controller that enables
a robot arm to move within an environment while regulating contact
forces across its entire surface. The controller uses model predictive
control (MPC) with a time horizon of length one and a linear
quasi-static mechanical model. At each time step, the controller
constructs a model and solves an associated quadratic programming
problem in order to minimize the predicted distance to a goal subject
to constraints on the predicted contact forces.

\begin{figure}[t!]
\centering
\vspace{0.2cm}
\includegraphics[height=4.3cm]{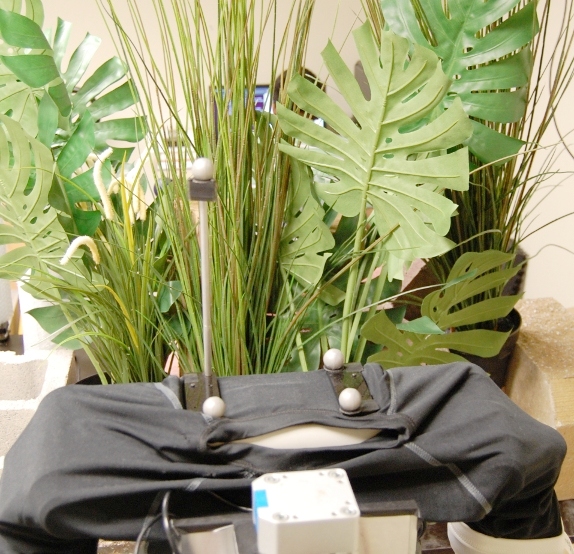}
\hspace{0.1cm}
\includegraphics[height=4.3cm]{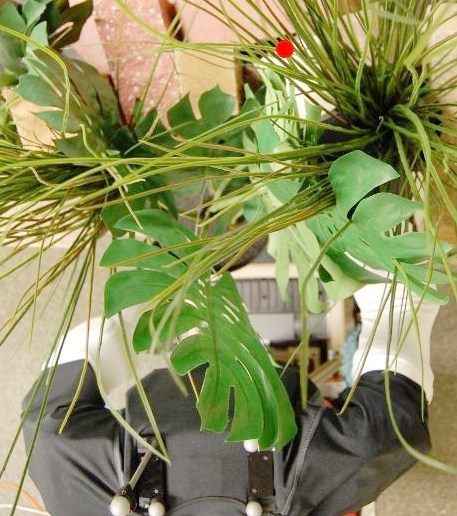}
\mycaption{
\label{fig:main_result}
\textbf{Left:} View of foliage from the robot's perspective. Two rigid
blocks of wood are occluded by the leaves. \textbf{Right:} Image of
the robot after it has successfully reached the goal location using
the controller we present in this paper. The red circle denotes the
position of the end effector.}
\end{figure}

We also empirically evaluate our controller's performance with respect
to the task of haptically reaching to a goal location in high clutter
(see \figref{fig:main_result}). We assume that the clutter can consist
of a variety of fixed, movable, and deformable objects, and that the
robot does not have a model of the environment in advance. This task
is representative of real-world challenges for robots, such as
retrieving objects from rubble, foliage, or the back of a shelf. It is
also representative of an animal reaching for food while foraging.

We tested our controller under a variety of conditions with a
simulated robot, a real robot with simulated tactile sensing, and a
real robot with real tactile sensors across its forearm. For many of
the tasks, the robots compressed, bent, or moved objects out of the
way with their arms while reaching the goal location. Our results
demonstrate that the model predictive controller has a higher success
rate and lower contact forces compared to a baseline controller.



\subsection{Biological Inspiration}
\label{sec:inspiration}

\begin{figure*}[t!]
\centering
\includegraphics[height=4.00cm]{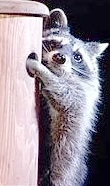}
\hspace{0.25cm}
\includegraphics[height=4.00cm]{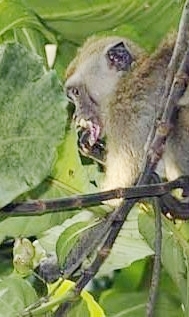}
\hspace{0.25cm}
\includegraphics[height=4.0cm]{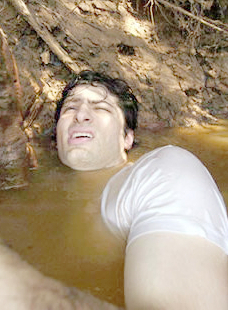}
\includegraphics[height=4.0cm]{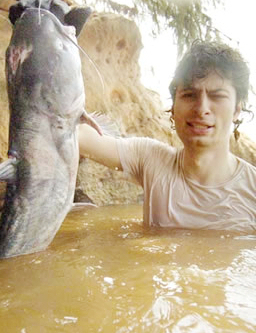}
\hspace{0.25cm}
\includegraphics[height=4.0cm]{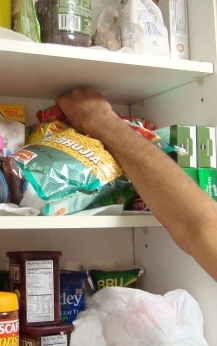}
\hspace{0.25cm}
\includegraphics[height=4.0cm]{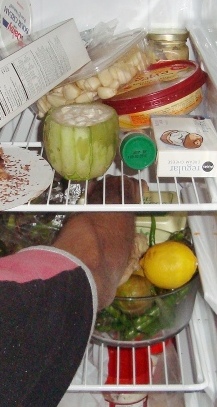}
\captionof{figure}{\textit{\label{fig:foraging} We propose
    foundational capabilities for robotic manipulation that will
    enable robots to make and exploit contact with their environment.
    While foraging for food, animals and humans make contact at
    multiple locations on their arm and operate in cluttered
    environments. (a) A raccoon reaches into a bird house to find eggs
    and young \citep{racoon_eggs}.  (b) A Long-tailed Macaque grasps
    fruit in dense foliage \citep{long_macaque}. (c) When noodling,
    people find catfish holes from which to pull fish out
    \citep{noodling}. (d)-(e) A person makes contact along his forearm
    while reaching for an object in the back of a shelf and
        refrigerator. (All images used with permission)}}
\vspace{-0.3cm}
\end{figure*}

Animals serve as an inspiration for our research (see
\figref{fig:foraging}), especially in terms of the capabilities they
exhibit, their sensing, and their actuation.

Animals dramatically outperform current autonomous robots within
unstructured environments, such as when foraging in dense
foliage. During these activities, animals often make contact with the
world at multiple locations along their arms, and reach into visually
occluded spaces.

Touch is an important sensory modality for successful foraging
\citep{dominy2004fruits, iwaniuk1999skilled}. In general, animals can
usefully manipulate the world in the absence of vision. As an extreme
example, the star-nosed mole uses the sense of touch almost
exclusively while foraging \citep{catania1999nose}. Humans also
competently manipulate the world without vision, as the reader can
demonstrate by haptically exploring the underside of a nearby
table. Inspired by these capabilities, our goal is to develop methods
that degrade gracefully when deprived of non-haptic modalities, such
as vision and audition.

Animals also serve as inspiration for our decision to assume the
presence of whole-body tactile sensing. Although whole-body tactile
sensing is currently rare in robotics, it is nearly ubiquitous in
biology, which suggests that it is advantageous for operation in
unstructured environments.  Organisms from small nematodes to insects
and mammals are able to sense forces across their entire bodies
\citep{mechanotransduction2007,wormbook2006,cell_bio_of_touch2010,klatzky2009}.
Sensing forces also plays an important role in avoiding injury. For
example, loss of sensitivity in a human diabetic's foot is a strong
risk factor for injuring the foot \citep{foot_injury}. As has often
been noted, tactile sensing also supports human manipulation
\citep{johansson2009coding}.


Compliant actuation at the joints is another common characteristic in
animals that we have chosen to emulate \citep{hogan1984adaptive,
  alexander1990three, migliore2005biologically}. Robotics researchers
have demonstrated that compliant joints lower interaction forces
during incidental contact and can be beneficial for
unmodeled and dynamic interactions \citep{pratt_vmc, pratt1995series,
  buerger2007complementary, low_impedance_walking_robots_pratt}. This
capability is now relatively common within robotics, although we use
stiffnesses that tend to be lower than other published research.  For
example, in some postures, the stiffness at the end effector is a
factor of five lower than those reported by DLR in
\cite{dlr_door}. The values we use are similar to measured stiffnesses
of humans during planar reaching motions \citep{shadmehr1993control}.


\subsection{Benefits of Whole-body Contact and Tactile Sensing}

\begin{figure}[!t]
\begin{center}
\includegraphics[height=3.2cm]{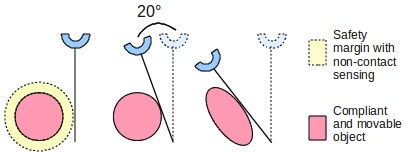}
\end{center}
\vspace{-0.2cm}
\mycaption{
\label{fig:one_dof_example}
Example illustrating the available range of motion for a 1 DoF arm
if the controller uses a safety margin with non-contact sensing
\textbf{(left)}, allows contact with an object \textbf{(middle)}, and
allows the arm to push into compliant and movable objects
\textbf{(right)}.}
\end{figure}

Given our emphasis on whole-body contact and whole-body tactile
sensing, we now illustrate some of the performance benefits associated
with these design decisions. 

One benefit of allowing contact with the arm is the increased
effective range of motion of the manipulator. As illustrated in
\figref{fig:one_dof_example}, the performance loss due to avoiding
contact is exacerbated by safety margins and an inability to apply
forces that compress or move objects. Similarly, if the robot has a
compliant exterior, avoiding contact forfeits the additional range of
motion achievable by compressing this exterior. Animals appear to use
this method to achieve greater motion while in contact and squeeze
through openings. The reader can gain some insight into this by noting
the compressibility of the human forearm that results from the soft
tissues surrounding the endoskeleton.

%

Whole-body tactile sensing with high spatial resolution also has
advantages in terms of distinguishing between distinct contact
configurations and force distributions, and measuring forces with high
sensitivity. Prior research has attempted to use the geometry of
links, measurements of joint torques and force-torque sensors to
estimate contact properties (e.g, \cite{eberman1990determination,
  bicchi1993contact, kaneko1994contact, de1999estimating}). However,
interpretation of data from these sensors can often be ambiguous in
multi-contact situations \citep{salisbury1984interpretation}. In
practice, the estimation can also be sensitive to the configuration of
the manipulator, the fidelity of the torque estimates, and friction
and flexibility at the joints \citep{dogar2010, eberman1989whole}.

\begin{figure}[t!]
\begin{center}
\includegraphics[height=2.0cm]{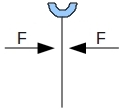}
\hspace{2.0cm}
\includegraphics[height=2.0cm]{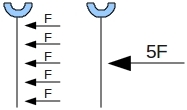}
\end{center}
\vspace{-0.2cm}
\mycaption{
\label{fig:multi_contact}
Some multi-contact conditions can not be detected \textbf{(left)} or
distinguished \textbf{(right)} using only joint torque sensing
or force-torque sensors mounted at the joints. We can detect and
distinguish between these conditions using tactile sensors covering
the arm.
}
\vspace{-0.3cm}
\end{figure}

\figref{fig:multi_contact} shows two examples of contact conditions
that will result in ambiguity if a robot only uses joint torque
sensing or force-torque sensors mounted at the joints. In the first
example, the resultant force and torque on the robot arm is zero, but
it is wedged between two contacts. The second example illustrates that
contact over a large area, and a high force at a single point can
result in the same total resultant force and torque.

Distinguishing among these situations can be advantageous. For
example, a high total force distributed over a small area, such as due
to contact with the edge of a cinder block or a small branch, has
greater potential to damage the robot or the world,
respectively. Similarly, the same large total force distributed across
a large area due to contact with tall grass or leaves is less likely
to damage the robot or the world. Moreover, the geometries associated
with distinct contact regions, such as a rigid point, line, or plane,
imply distinct options for subsequent movement.

More generally, for many manipulation tasks, the manipulator primarily
influences the world via contact forces with other forms of physical
interaction, such as heat transfer, being uncontrolled or irrelevant.
As such, we expect that direct measurement of contact forces will
enable superior manipulation capabilities.


\subsection{Challenges Associated with Reaching in High Clutter}
\label{sec:current_approaches}

For this paper, we focus on the task of reaching to a goal location in
high clutter. This entails a number of challenges, including the
following:

\begin{itemize}
\item \textit{Lack of non-contact trajectories:} As clutter increases,
  approaches that avoid contact with the environment will have a
  diminishing set of trajectories that can successfully perform the
  task. If the robot is interacting with movable or compliant objects,
  such as foliage, reaching the goal while applying low forces might
  be possible but non-contact trajectories may not exist.
\item \textit{Contact with only the end effector may be inefficient or
  infeasible:} Removing or rearranging the clutter by making serial
  contact with only the end effector may be inefficient or
  infeasible. For example, given an environment with multiple
  compliant objects (e.g., plants) it may not be possible to first
  bend each of the objects out of the way one at a time without
  plastic deformation. Instead, an efficient solution would be to
  directly reach to the goal and allow multiple contacts to occur with
  the compliant objects and the arm.
\item \textit{Clutter can consist of unique objects and configurations
  that have not been encountered before:} For some types of natural
  clutter, such as dense foliage, each object encountered can be
  unique. Objects could be fixed, movable, rigid, deformable,
  granular, fluid-like, and dynamic. Likewise, the configuration of
  the environment can be unique. Statistical properties may be
  informative, but a specific environment may only be encountered once
  by the robot.
\item \textit{Observation of geometry is obstructed:} Low visibility
  due to occlusion will often prevent conventional line of sight
  sensors, such as cameras and laser range finders, from modeling the
  geometry of the clutter in advance (see \figref{fig:main_result}).
\item \textit{Mechanics are difficult to infer without contact:}
  Non-contact sensing provides limited ability to infer the mechanical
  properties of the clutter, such as whether or not an object can be
  bent or moved out of the way. Likewise, objects may be mechanically
  coupled in complicated ways, such as through adhesion or unobserved
  rigid connections.
\end{itemize}

Notably, many approaches to manipulation are poorly matched to address
these challenges. For example, approaches that rely on preexisting
detailed models, estimation of models via conventional line-of-sight
sensing, or collision-free motions with the arm would fare poorly
under real-world conditions at which animals excel (e.g.,
\cite{planning_springer_handbook, stilman2007manipulation,
  srinivasa2009herb, saxena2008rgn}).




\subsection{Our Approach}
\label{sec:our_approach}

In contrast, our approach directly addresses these challenges
associated with high clutter due to the following properties:

\begin{itemize}
\item \textit{We explicitly allow multiple contacts across the entire
  surface of the arm:} Our approach assumes that the entire surface of
  the manipulator is covered with pressure sensing elements (tactile
  pixels or taxels), and that every taxel could be simultaneously in
  contact with the world at any given moment.
\item \textit{We do not require a detailed model of the environment
  prior to contact:} Our approach only requires that initial
  parameters appropriate for the robot, the environment, and the task
  be provided to the robot's control system in advance. For our
  current controller, this includes the force magnitude below which no
  penalty is expected, and the initial stiffness estimate assigned to
  new contacts.
\item \textit{We only require contact-based sensing:} Our approach
  only requires contact-based sensing. Our current controller only
  makes use of haptic sensing in the form of joint angles\footnote{Due
    to the low-stiffness virtual visco-elastic springs at the robot's
    joints, the joint angles over time directly relate to the joint
    torques.} and taxel responses. Through this contact-based sensing
  and a kinematic model of its own arm, the controller estimates the
  instantaneous contact geometry and the stiffness associated with
  each contact in order to generate a local model. Future work may
  make use of non-contact sensing, which would be complementary, but
  this is not required. 
\end{itemize}


\subsection{Organization of this Paper}
\label{sec:organization}
The rest of this paper is organized as follows. In
\secref{sec:related_work} we discuss related research on manipulation
in clutter, multi-contact manipulation, motion planning with
deformable objects, robot locomotion, and model predictive control.
Next, in \secref{sec:controller}, we derive our model predictive
controller for the task of reaching to a goal location in a cluttered
environment.  We describe the hierarchy of controllers running on our
robot, the linear quasi-static model that the controller uses, and the
quadratic program that the controller solves at each time step.

We then describe three testbeds that we used to empirically test the
performance of our controller (\secref{sec:testbeds}) and the baseline
controller that we compared our model predictive controller against
(\secref{sec:approach_compare}). Next, we describe the experiments
that we ran in \secref{sec:experiments}. We end with a discussion of
our work in \secref{sec:discussion} and conclude with a brief summary
of our results in \secref{sec:conclusion}.

\section{Related Work}
\label{sec:related_work}

\subsection{Manipulation in Clutter}

Within this paper, our goal is to enable robots to reach to a goal
location in cluttered environments and manipulate with multiple
contacts across the entire arm using haptic sensing.

In contrast, robotics research has often addressed the task of
generating collision free trajectories (e.g, \cite{lozano1987simple,
lavalle2001randomized, planning_springer_handbook}), generating
reaching motions in free space (e.g, \cite{mettaforce,
stulp2009compact, hersch2006biologically}), and manipulating objects
in uncluttered environments (e.g, \cite{natale2006sensitive, jain2010auro,
hsiao2010contact, pastor2011online, romano2011human, saxena2008rgn}).

Research has also looked at the problem of manipulation in cluttered
environments. However, most prior research on manipulation in clutter
with autonomous control and during teleoperation (e.g,
\cite{leeper2011strategies}) restricts contact between the robot and
its environment to the end effector. Often, prior research has also
used non-contact, line of sight sensors and required pre-existing
models of objects.

\cite{stilman2007manipulation} describes an algorithm for planning in
an environment with movable obstacles. Within software simulation, the
planner uses geometric models of all the objects in the world to
enable a robot to rearrange clutter by grasping and moving objects,
and opening doors.

\cite{dogar2011fpc} presents a framework to plan a sequence of actions
such as pushing and grasping objects to rearrange clutter prior to
grasping an object. The actions currently used within the framework
restrict contact to the robot's end effector and avoid other contact
with the world. The implementation relies on estimating the pose of
objects in the environment using visual and geometric models (created
during an off-line modeling stage). Additionally, for now, the planned
actions are executed without sensor feedback.

\cite{mason2011autonomous} describes a simple end effector design that
can be used to grasp a single marker from a cluttered pile of nearly
identical markers, and haptically estimate the marker's pose after
grasping it. This is the most similar work in spirit to ours, since it
investigates manipulation in high clutter, does not use a detailed
model of the environment prior to contact, allows multiple contacts
across the surface of the end effector, and uses haptic sensing. Their
``Let the fingers fall where they may.'' approach to grasping in
clutter has other notable similarities to our approach to reaching in
clutter. Both approaches use greedy controllers that are run
iteratively, and both approaches ignore the details of how the clutter
responds to the robot's actions. However, our approach performs more
complex feedback-based control of the manipulator and does not use a
simple mechanism nor simple sensing, which they emphasize. This is in
part because we wish to regulate the contact forces. More degrees of
freedom also appear to improve the performance of our controller,
although we do not yet have reportable results to support this
claim. In addition, we focus on reaching in clutter, rather than
grasping, and present empirical results for diverse environments, in
contrast to their experiments with a collection of durable, rigid,
nearly-identical, manufactured objects.

%


\subsection{Multi-contact Manipulation}

\cite{Park2008} presents a framework for controlling a robot
with multiple contacts along the links. It generalizes previous
direct force control methods \citep{raibert1981hybrid,
khatib1987unified} to not require force and motion to be along
orthogonal directions in Cartesian space and to allow for contacts at
points other than the end effector.

This method requires a full dynamic model of the robot and assumes
stationary and rigid contacts. Further, this framework assumes that
the robot has at least six degrees of freedom (DoF) for each contact,
to control the contact force and torque vector \citep{Sentis2010}. A
seven degree of freedom arm, like the robot arm that we use, with
multiple contacts is unlikely to have six independent degrees of
freedom for each contact.


Using this framework, results have been shown in simulation
\citep{sentis2005swb, Sentis2010}, and on a real robot in relatively
controlled settings \citep{petrovskaya2007}. No results have been
shown in cases where the robot makes additional unpredicted contact
with the environment or loses contact at some locations.

In contrast, our controller uses a linear quasi-static model of the
robot's interaction with the environment and does not assume that the
robot has six degrees of freedom for each contact.  On a real robot
with a tactile skin sensor, we demonstrate that our controller can
operate in cluttered environments with multiple unpredicted contacts
with compliant, rigid, movable, and fixed obstacles across the entire
arm of the robot.

Research in motion planning for humanoid robots has shown that a robot 
with geometric models of its environment can make contact at multiple,
predetermined locations on its body to better perform a task, such as
lean on a table to take a large foot step \citep{Legagne2011}, use
contacts at hands and knees to climb a ladder in simulation
\citep{hauser2005non}, and use contact between the hand and a table
while sitting down \citep{escande2009contact}. These approaches
require a complete geometric model of the world (which can be
difficult or impossible to obtain unless the robot is operating in a
controlled environment), assume stationary and rigid contacts, and do
not incorporate sensor feedback as the robot (real or simulated)
executes the planned kinematic trajectory. These approaches attempt to
maintain balance on the humanoid robot while we assume that the robot
is statically stable.

There is also research on multi-contact manipulation within the
context of using all the surfaces of a multi-fingered hand, or the
entire body to grasp and manipulate a single object (e.g,
\cite{bicchi1993force, bicchi2000robotic, hsiao2006imitation,
platt2003extending}).

\subsection{Motion Planning with Deformable Objects}
Manipulation research often assumes that objects that the robot
interacts with are rigid. At the same time, there is research on
motion planners that allow the robot to make contact with, and push
into deformable objects (e.g.  \cite{frank2011using,
rodriguez2006planning, patil2011motion}). However, these approaches
assume knowledge of the specific configuration of the objects and
require accurate and detailed models of how objects deform. We avoid
these assumptions in our work.

These approaches build object deformation models by using data-driven
methods for a specific object, or computationally expensive physics
simulations that use the physical properties of the objects. Accurate
object deformation models can be hard to obtain in realistic and
cluttered environments. Additionally, if multiple objects are in
contact with each other and the specific configuration is unknown,
then building these models before making contact may not be feasible.

\subsection{Robot Locomotion}
Our approach to robot manipulation has similarities to approaches that
have been successful for robot locomotion. For example, researchers
have developed robots that locomote in cluttered environments without
detailed geometric models of the terrain nor planning over long time
horizons \citep{saranli2001rhex, raibert2008bigdog}. Likewise, whole
body contact, and contact in general, has not been considered
undesirable. For example, robots have used contact all over their
bodies to traverse the ground and swim in granular media
\citep{mckenna2008toroidal, maladen2010biophysically}. Additionally,
the use of simple mechanical models, compliance, and force sensing is
common for robot locomotion \citep{raibert2008bigdog,
low_impedance_walking_robots_pratt, pratt_vmc, garcia1998simplest}.

\subsection{Model Predictive Control}

One of the initial application areas for model predictive control
(MPC) was chemical process control \citep{garcia1989model}. It is
often referred to as receding horizon control when used for control of
aerial vehicles \citep{bellingham2002receding, abbeel2010autonomous}.
MPC has also been used in research in robot locomotion (e.g,
\cite{wieber2006trajectory, manchester2011stable, erezinfinite}), and
for controlling robot manipulators (e.g, \cite{ivaldi2010approximate,
from2011motion, kulchenko2011first}). 


\section{Model Predictive Controller}
\label{sec:controller}

The controller that we have developed uses linear model predictive
control (MPC) with a time horizon of length one.  Specifically, using
the notation of \cite{morari1999model}, our controller uses a linear
discrete time model of the system,
\begin{align}
x(k+1) &= Ax(k) + Bu(k),
\label{eqn:mpc_morari}
\end{align}
where $x(k)$ is the state of the system and $u(k)$ is the control
input.

At each time step, $k$, the controller computes a sequence of control
inputs, $u^{*}(i), i=k \dots (k+N-1)$, to minimize a quadratic
objective function of $x(k), \dots, x(k+N)$ and $u(k), \dots,
u(k+N-1)$, subject to linear inequality constraints on $x(k), \dots,
x(k+N)$ and $u(k), \dots, u(k+N-1)$, where $N$ is the length of the
time horizon of the model predictive controller. This defines a
quadratic program \citep{morari1999model}.  The controller then uses
only the first control input, i.e. it sets $u(k) = u^{*}(k)$, and
reformulates the quadratic program at the next time step. In this
paper, we use a time horizon of length one $(N = 1)$, and recompute
the $A$ and $B$ matrices in \eqnref{eqn:mpc_morari} at each time step.

In the rest of this section, we describe our model predictive controller
for manipulation with multiple contacts. First, in
\secref{sec:qp_summary}, we give an overview of the controller that we
have developed. Next, we present the hierarchy of controllers running
on our robot in \secref{sec:control_structure}. In
\secref{sec:linear_model} we describe the linear quasi-static model
that our model predictive controller uses, and detail the quadratic
program that we solve at each time step \secref{sec:set_up_qp}. We
then describe some extensions to the quadratic program in
\secref{sec:qp_extensions}.

\subsection{Overview of the One-Step Model Predictive Controller}
\label{sec:qp_summary}
The model predictive controller that we have developed uses a linear
discrete time model of the system, a one step time horizon, and
attempts to move the end effector along a straight line to the goal
subject to constraints on the predicted contact forces.

It explicitly allows the robot to apply any force less than a
\mytexttt{don't care force threshold} at each contact. Our controller
has the following parameters that influence its behavior:
\begin{itemize}
\item \textit{Goal location $(x_g \in\Re^3)$:} This is the location
that the controller attempts to move the end effector to.
\item \textit{Contact stiffness matrices $(K_{c_i}\in\Re^{3\cross
3})$:} These are the controller's estimates of the stiffness matrices
for each contact location along the arm. In this paper, we assume that
the stiffness at each contact is non-zero along the direction normal
to the surface of the robot arm and is zero in the other directions. 
\item \textit{Don't care force thresholds $(f_{c_i}^{thresh}
\in\Re^3)$:} The controller attempts to keep the force at
each contact below this value, and applies no penalty to contact
forces below this threshold.
\item \textit{Maximum rate of change of contact force $(\Delta
f_{c_i}^{rate}\in\Re^3)$:} This term limits the predicted change in
the contact force over one time step with the goal of preventing large
and abrupt changes in the contact force.
\item \textit{Safety force threshold $(f_{c_i}^{safety} \in\Re^3)$:}
If the contact force, $f_{c_i}$, exceeds this safety threshold value,
the controller stops updating the virtual trajectory and we report it
as a failure of the controller.
\end{itemize}

In this paper, we perform experiments on three different testbeds,
described in \secref{sec:testbeds}. The precise meaning of contact,
and thus the \mytexttt{don't care force threshold} and other
parameters of the model predictive controller, depends on the specific
testbed, as described in \secref{sec:tactile_feedback}.

\begin{figure}
\begin{tikzpicture}[auto, node distance=1cm,>=latex']
    \node [nooutlineblock] (input) {\small Goal Location};
    \node [clearblock, below of=input, node distance=1.5cm] (qp) {\small Model
        Predictive Controller};

    \node [clearblock, below of=qp, node distance=1.2cm] (sum)
{$\phi = \phi_{k-1} + \Delta \phi^*$};

    \node [clearblock, below of=sum, node distance=1.8cm] (simple)
          {$\tau =  K_j (\phi - \theta) + D_j \dot\theta +\hat{\tau}_g$};
    \linespread{1.0}
    \node [nooutlineblock, right of=simple, node distance=3.3cm, text
    width=2.5cm] (simpletext) {\begin{center} \scriptsize ``Simple''
        Impedance Control \citep{hogan2005impedance} \end{center}};

    \node [clearblock, below of=simple,
            node distance=1.8cm] (plant) {$M\ddot{\theta} +
                C\dot{\theta} + \sum_{i=1}^n
                    J_{c_i}^T f_{c_i} + \tau_g = \tau$};
    \linespread{1.0}
    \centering
    \node [nooutlineblock, right of=plant, text width=3cm,
           node distance=4.5cm] (planttext) {\scriptsize Plant};

    \linespread{1.0}
    \node [clearblock, left of=simple, node distance=3.5cm, text width=1.0cm]
    (measurements) {\scriptsize\begin{center}Joint Encoders\end{center}};

    \node [clearblock, left of=sum, node distance= 3.5cm, text width=2.2cm]
    (qp_feedback) {\scriptsize\begin{center}Joint Encoders, Tactile Skin
        \end{center}};


    \draw [myarrow] (input) -- node [near start] {$x_g \in \Re^3$} (qp);
    \draw [myarrow] (qp) -- node {$\Delta \phi^{*}$} (sum);
    \draw [myarrow] (simple) -- node [near start] {$\tau$} (plant);
    \draw [myarrow] (sum) -- node [near start] {$\phi$} (simple);
    \draw [myarrow_notrounded] (plant) -| (measurements);
    \draw [myarrow] (measurements) -- node {$\theta, \dot{\theta}$} (simple);

    \node[coordinate, left of=measurements, name=waypoint1] {};
    \draw [myline] (plant) -| node {} (waypoint1);
    \node[coordinate, above of=waypoint1, name=waypoint2, node
    distance=0.4cm] {};
    \draw [myarrow] (waypoint1) -- node {} (qp_feedback.south);

    \draw [myarrow] (qp_feedback) |- node {$\theta, f_{c_i}, J_{c_i}$} (qp);

    \draw [myline, dashed] (-4.7,-0.8) -- (4.3,-0.8);
    \draw [myline, dashed] (-4.7,-3.6) -- (4.3,-3.6);
    \draw [myline, dashed] (-4.7,-5.4) -- (4.3,-5.4);

    \node [nooutlineblock] at (3.3, -1.5) (text) {\normalsize \textit{50-100Hz}};
    \node [nooutlineblock] at (3.3, -3.9) (text) {\normalsize \textit{1kHz}};

\end{tikzpicture}
\mycaption{\label{fig:block_diagram}Block diagram showing the
hierarchical control structure and the equations of motion. Details
are in \secref{sec:controller}. The controller frequencies are
specific to our implementation.}
\end{figure}

\subsection{Control Structure}
\label{sec:control_structure}

In this work, we use a hierarchical control structure with an inner
1kHz real time joint space impedance controller, termed ``simple''
impedance control by \cite{hogan2005impedance}, and an outer model
predictive controller that runs at 50-100Hz, as shown in
\figref{fig:block_diagram}. 

%

Researchers have argued for the benefits of robots with low mechanical
impedance \citep{low_impedance_walking_robots_pratt,
buerger2006stable}. As has often been noted, these arguments are
particularly relevant for manipulation in unstructured environments,
since robots are likely to be uncertain about the state of the world.
At minimum, low impedance can reduce the forces and moments resulting
from unpredicted contact, and thus reduce the risk of damage to the
robot, environment, and nearby people.

\mysubsubsection{``Simple'' Impedance Control}
\label{sec:epc}

For a detailed description and analysis of this form of impedance
control, we refer the reader to \cite{hogan2005impedance}. The input
to the  1kHz ``simple'' impedance controller, $\phi$, is called a
virtual trajectory. The controller uses feedback from the joint
encoders to command torques at the joints, $\tau$, that are given by
\begin{align}
\tau & =  K_j (\phi - \theta) + D_j \dot\theta + \hat{\tau}_g(\theta).
    \label{eqn:virtual_visco}
\end{align}
$K_j$ and $D_j$ are constant $m\cross m$ diagonal joint-space
stiffness and damping matrices, $\theta\in\Re^m$ and
$\dot{\theta}\in\Re^m$ are the current joint angles and joint
velocities, and $\hat{\tau}_g\in\Re^m$ is a gravity compensating
torque vector which is a function of $\theta$. The robot arm has $m$
joints.

As a result, the closed loop system behaves as if the arm is
connected to the joint-space virtual trajectory, $\phi$, via torsional
visco-elastic springs at the joints.  If $\phi$ is held constant,
``simple'' impedance control can be shown to result in stable
interaction with passive environments for contacts all over the arm
\citep{hogan2005impedance, hogan1988stability}.

Unlike other approaches to force control and impedance control,
``simple'' impedance control does not explicitly model the dynamics of
the arm nor the impedance at the end effector \citep{Sentis2010,
albu2003cartesian}.  We have found in our previous work that in
practice this form of impedance control, also referred to as
``equilibrium point control'', allows the robot to interact
with the world in a stable, compliant, and effective way \citep
{kemp_roman_2007, kemp_icar_2007, jain2009bbd, jain2009pon,
jain2010pod}.

Other researchers have looked at similar robotic control strategies in
simulation \citep{gu2006epb}, in free-space motions
\citep{williamson1996ppi}, in legged locomotion \citep{migliore2009},
and in rhythmic manipulation from a fixed based
\citep{williamson1999rac}.

\mysubsubsection{Model Predictive Controller}

The model predictive controller is part of the outer feedback control
loop that runs between 50-100Hz in our implementation, as shown in
\figref{fig:block_diagram}. The input is a goal location, $x_g \in
\Re^3$, that the controller attempts to reach. The controller uses
feedback from the joint encoders and the tactile skin to compute
$\Delta\phi^* \in\Re^m$, an incremental change in the virtual
joint-space trajectory. This $\Delta\phi^*$ is the control input,
$u(k)$, of \eqnref{eqn:mpc_morari}.

We will now derive the model predictive controller.

%

\subsection{Linear Discrete-Time Model}
\label{sec:linear_model}

\begin{figure}[t]
\begin{center}
\includegraphics[height=6.0cm]{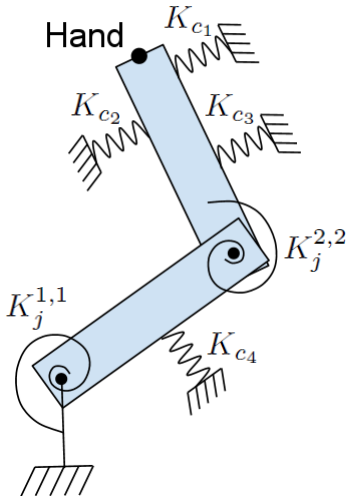}
\end{center}
\mycaption{
\label{fig:mech_model}
Graphical representation of a planar version of the quasi-static
mechanical model with torsional springs at the joints of the robot and
linear springs at contacts that our model predictive controller uses,
as described in \secref{sec:linear_model}.}
\end{figure}

In this section, we derive a discrete time linear quasi-static model,
similar to \eqnref{eqn:mpc_morari}, for the arm and its interaction
with the world that our model predictive controller uses.
Specifically, the model will be of the form
\begin{align}
\theta(k+1) &= \theta(k) + B \Delta\phi(k),
\label{eqn:mpc_model}
\end{align}
where $\theta \in \Re^m$ is the state of the system (vector of joint
angles for a robot with $m$ joints), the control input
$\Delta\phi\in \Re^m$ is the incremental change in the
joint-space virtual trajectory of the impedance controller, and
$B \in\Re^{m\cross m}$.

We begin by assuming that the robot has a fixed and statically stable
mobile base and the arm is in contact with the world at $n$ locations.
We denote the $i^{th}$ contact as $c_i$. The equations of motion in
joint space are
\begin{align}
M(\theta)\ddot{\theta} + C(\theta, \dot\theta)\dot{\theta} +
\sum_{i=1}^n J_{c_i}^T(\theta) f_{c_i} + \tau_g(\theta) &= \tau,
\label{eqn:eqn_of_motion}
\end{align}
where $f_{c_i}\in\Re^3$ is the force at the $i^{th}$ contact, 
$J_{c_i}\in\Re^{3\cross m}$ is the Jacobian matrix for contact $c_i$,
$\tau_g \in\Re^m$ is the vector of torques due to gravity at each
joint, and $\tau \in\Re^m$ is the vector of torques applied by the
actuators at the joints. \eqnref{eqn:eqn_of_motion} ignores effects
such as friction at the joints, but is commonly used in robotics
\citep{dynamics_springer_handbook}.

Combining the equations of motion (\eqnref{eqn:eqn_of_motion}) with
the impedance control law (\eqnref{eqn:virtual_visco}) gives us the
model of the arm and its interaction with the world as
\begin{align}
M\ddot{\theta} + C\dot{\theta} +
\sum_{i=1}^n J_{c_i}^T f_{c_i} + \tau_g &= K_j (\phi - \theta) + D_j
\dot\theta + \hat{\tau}_g.
\label{eqn:full_model}
\end{align}
In this paper, as an approximation, we assume that the dynamics are
negligible, and that the gravity compensating torques are perfect.
So, we remove all terms with $\ddot\theta$ or $\dot\theta$ from
\eqnref{eqn:full_model}, and set $\hat{\tau}_g = \tau_g$ to get
\begin{align}
\sum_{i=1}^n J_{c_i}^T f_{c_i} &= K_j (\phi - \theta),
\label{eqn:simple_model}
\end{align}
which is a quasi-static model. In \eqnref{eqn:simple_model}, the
torques at the joints due to the contact forces (left-hand side)
balance the torques applied by the actuators in the joints (right-hand
side).


For the contact model, we ignore friction at the contacts and assume
that each contact behaves like a linear spring with the contact force
along the normal vector of the surface of the robot arm. These
assumptions are similar to the Hertzian contact model
\citep{contact_springer_handbook, hertz_contact_mechanics_book}. This
results in a mechanical model with torsional springs at the joints
and linear springs at the contacts, shown in \figref{fig:mech_model}.


If we take the difference of \eqnref{eqn:simple_model} at time
instants $k$ and $k+1$, we get
\begin{align}
\sum_{i=1}^n J_{c_i}^T(k+1)& f_{c_i}(k+1) - J_{c_i}^T(k) f_{c_i}(k) =
\nonumber
\\
&K_j (\phi(k+1) - \phi(k) - \theta(k+1) + \theta(k)).
\label{eqn:simple_diff_1}
\end{align}
We assume that the change in the configuration of the arm in one time
step, $\theta(k+1) - \theta(k)$, is small and we approximate
$J_{c_i}(k+1)$ with $J_{c_i}(k)$. This reduces \eqnref{eqn:simple_diff_1} to
\begin{align}
\sum_{i=1}^n J_{c_i}^T(k)& (f_{c_i}(k+1) - f_{c_i}(k)) = \nonumber \\
&K_j (\Delta \phi(k) - \theta(k+1) + \theta(k)),
\label{eqn:simple_diff_2}
\end{align}
where $\Delta\phi(k) = \phi(k+1) - \phi(k)$ is the control input of
the model predictive controller, see \eqnref{eqn:mpc_model} and
\figref{fig:block_diagram}.

Using the linear elastic spring model for the contacts, shown in
\figref{fig:mech_model},
\begin{align}
f_{c_i}(k+1) - f_{c_i}(k) &= K_{c_i} J_{c_i} \Delta\theta(k),
\label{eqn:delta_fc}
\end{align}
where $\Delta\theta(k) = \theta(k+1) - \theta(k)$.
We can now use \eqnref{eqn:delta_fc} to rewrite
\eqnref{eqn:simple_diff_2} as
\begin{align}
\theta(k+1) &= \theta(k) +
\left(K_j + \sum_{i=1}^n J_{c_i}^T K_{c_i} J_{c_i} \right)^{-1} K_j \Delta \phi(k).
\label{eqn:mpc_model_final}
\end{align}
$\left(K_j + \sum_{i=1}^n J_{c_i}^T K_{c_i} J_{c_i}
\right)$ is the sum of a positive definite matrix, $K_j$, and positive
semi-definite matrices, $J_{c_i}^T K_{c_i} J_{c_i}$, and is therefore
positive definite and invertible.

\eqnref{eqn:mpc_model_final} is in the same form as Eqns.
\ref{eqn:mpc_morari} and \ref{eqn:mpc_model}. This is the linear
discrete time model of the system that our controller generates and
uses at each time step. We use contact forces and locations from
whole-arm tactile sensing, and joint angles from encoders at the
joints to estimate $f_{c_i}$, $J_{c_i}$, and $K_{c_i}$.

The linear form of \eqnref{eqn:mpc_model_final} allows us to frame the
optimization as a quadratic program, which can be solved efficiently
(\secref{sec:set_up_qp}).  Additionally, we empirically demonstrate in
\secref{sec:experiments} that our controller performs well in the task
of reaching to a goal location in cluttered environments.

\subsection{Quadratic Program to Compute $\Delta\phi^*$}
\label{sec:set_up_qp}


In this section we describe the quadratic program (QP) that our model
predictive controller solves at each time step.

Specifically, using the terminology of \cite{boyd2004convex}, our
optimization variable is $\Delta\phi$, an incremental change in the
joint-space virtual trajectory, and we minimize a quadratic objective
function subject to linear equality and inequality constraints. We use
the open source OpenOpt framework to solve the quadratic program
\citep{openopt}.

In this paper, the objective function is of the form
\begin{align}
\sum_i \alpha_i g_i,
\label{eqn:objective}
\end{align}
where $g_i$ are quadratic functions of the optimization variable
$\Delta\phi$, and $\alpha_i$ are empirically tuned scalar weights.

We set up the quadratic program such that the solution,
$\Delta\phi^*$, will result in the predicted position of the end
effector which is closest to a desired position subject to constraints
on the predicted change in the joint angles and contact forces.

\mysubsubsection{Move to a Desired Position}
The first term of quadratic objective function of
\eqnref{eqn:objective} attempts to move the end effector to a desired
position. It is of the form
\begin{align}
g_1 &= \left\| \Delta x_{d}-\Delta x_{h}\right\| ^{2},
\label{eqn:objective_function}
\end{align}
where $\Delta x_h = x_h(k+1) - x_h(k)$ is the predicted motion of the
end effector (or hand) and $\Delta x_d \in\Re^3$ is the desired change in the
end effector position in one time step. In this paper, we attempt to
move the end effector in a straight line towards the goal, $x_g
\in\Re^3$, and compute
\begin{align}
\Delta x_d &= 
\begin{cases}
d_{w}\frac{x_g-x_h}{\left\| x_{g}-x_{h} \right\|} & \; if\:\left\|
    x_g-x_h\right\| > d_w \\
x_g-x_h & \; if\:\left\|
x_g-x_h \right\| \leq d_{w}\end{cases},
\label{eqn:delta_x_d}
\end{align}
where $d_w$ is a small constant distance. We use the kinematic relationship
\begin{align}
\Delta x_h &= J_h \Delta\theta,
\end{align}
where $J_h\in\Re^{3\cross m}$ is the Jacobian at the end effector (or
hand), and $\Delta\theta = \theta(k+1) - \theta(k)$ is the 
change in the joint angles predicted by the linear quasi-static
discrete time system model of \eqnref{eqn:mpc_model_final}. We can now
express the objective function $g_1$ as a quadratic function of
$\Delta\phi$:
\begin{align}
g_1 &= \left\| \Delta x_d - J_h \left(K_j + \sum_{i=1}^n J_{c_i}^T
        K_{c_i} J_{c_i} \right)^{-1} K_j \Delta\phi  \right\|^2.
\label{eqn:g1}
\end{align}


\mysubsubsection{Joint Limits}
We also add two linear inequality constraints to keep the predicted
joint angles within the  physical joint limits. These are of the form
\begin{align}
\Delta\theta_{min} \leq \Delta\theta \leq \Delta\theta_{max},
\label{eqn:joint_limit_constraint}
\end{align}
where $\Delta\theta_{min}$ and $\Delta\theta_{max}$ are the difference
between the minimum and maximum joint limits and the current
configuration of the robot.  Using
\eqnref{eqn:mpc_model_final} we can rewrite the
inequalities of \eqnref{eqn:joint_limit_constraint} as linear
inequalities in $\Delta\phi$.

\mysubsubsection{Contact Forces}
\label{sec:contact_forces}

For each contact, we attempt to restrict the contact force $f_{c_i}$ to
be below a \mytexttt{don't care force threshold} $f^{thresh}_{c_i}$
and limit the predicted change of the contact force, $\Delta f_{c_i} =
f_{c_i}(k+1) - f_{c_i}(k)$, in one time step. 
This results in two inequalities for each contact,
\begin{align}
\Delta f_{min} & \leq \Delta f_{c_i} \leq \Delta f_{max} \text{, where}
\label{eqn:force_constraints} \\
\Delta f_{min} & = -f_{c_i}^{rate} \text{, and} \\
\Delta f_{max} & = min\left(f_{c_i}^{rate}, f_{c_i}^{thresh} -
        f_{c_i} \right). \label{eqn:d_f_max}
\end{align}
$f_{c_i}^{rate}$ is a threshold on the maximum allowed
predicted change in the contact force in one time step.
The term $(f_{c_i}^{thresh} - f_{c_i})$ in \eqnref{eqn:d_f_max}
explicitly allows contact forces below $f_{c_i}^{thresh}$ without any
additional cost.

From Eqns. \ref{eqn:delta_fc} and \ref{eqn:mpc_model_final}, the
inequalities of \eqnref{eqn:force_constraints} can be expressed as
linear inequalities in $\Delta\phi$.

\subsection{Extensions to the Quadratic Program}
\label{sec:qp_extensions}
In this section, we describe three extensions to the quadratic program
of the previous section (\secref{sec:set_up_qp}) that we use in the
experiments of \secref{sec:experiments}.

\mysubsubsection{Squared Magnitude of $\Delta\tau$}
To discourage large changes in the joint torques in one time step, we
add a term 
\begin{align}
g_2 &= \left\|\Delta\tau\right\|^2 \nonumber \\
&= \Delta\phi^TK_j^TK_j\Delta\phi,
\end{align}
to the objective function after multiplying it with a scalar weight
$\alpha_2$, see \eqnref{eqn:objective}. This term is useful in
preventing large motions of the redundant degrees of freedom.

\mysubsubsection{Decrease Contact Forces Above Don't Care Threshold}

Due to modeling errors and unmodeled dynamics, the force at some
contact (or a number of contacts) can go above the \mytexttt{don't care
force threshold} $(f_{c_i} > f_{c_i}^{thresh})$. In this case, we
modify the inequality constraints of
\secref{sec:contact_forces} for these contacts to prevent an increase
in the predicted force. We also add an additional term $g_3$ to the
objective function that encourages the controller to decrease the
forces at these contacts.  This is of the form
\begin{align}
g_3 &= \sum_{i} \left\| \Delta f^d_{c_i}-\Delta f_{c_i}\right\| ^{2}
\text{ if } f_{c_i} > f_{c_i}^{thresh},
\label{eqn:force_objective_term}
\end{align}
where $\Delta f_{c_i}^d$ is the desired change the contact force in
one time step and $\Delta f_{c_i}$ is the change in the contact force
as predicted by the linear model that our controller uses.
We set $\Delta f_{c_i}^d$ as a force with a constant magnitude and a
direction opposite to $f_{c_i}$.  Using Eqns.  \ref{eqn:delta_fc} and
\ref{eqn:mpc_model_final}, we can express $g_3$ as a quadratic
function of $\Delta\phi$.


\mysubsubsection{Limits on the Virtual Trajectory}

On the robot Cody, described in \secref{sec:robot}, the joint-space
impedance controller limits the virtual trajectory to be within the
physical joint limits. To account for this, we add two additional
linear constraints on
$\Delta\phi$:
\begin{align}
\Delta\phi_{min}\leq\Delta\phi\leq\Delta\phi_{max}.
\end{align}

\section{Experimental Testbeds}
\label{sec:testbeds}

We evaluated our model predictive controller using three different
testbeds: 1) a software simulation testbed with a 3 DoF planar arm, 2)
a hardware-in-the-loop skin simulation testbed with a real 7 DoF arm,
and 3) a skin sensor covering the forearm of a real 7 DoF arm.  The same MPC
code written in Python runs on all three experimental testbeds.  For
visualization, we use the \texttt{rviz} program which is part of the
Robot Operating System \citep{quigley09}.


\subsection{Software Simulation}
\label{sec:software_testbed}

\begin{figure}
\centering
\includegraphics[height=4.0cm]{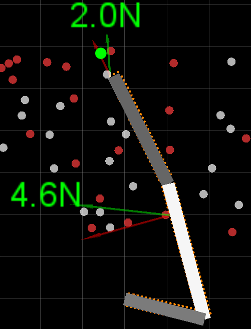}
\hspace{0.5cm}
\includegraphics[height=4.0cm]{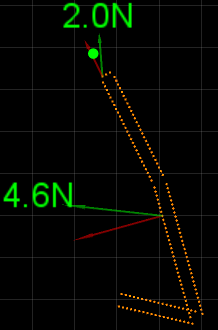}
\mycaption{
\label{fig:software_simulation}
\textbf{Left:} Visualization of the three link planar arm with tactile
skin interacting with obstacles within the software simulation
testbed.  Red obstacles are rigid and fixed, while gray obstacles are
rigid and movable. \textbf{Right:} Visualization of the whole-arm
tactile skin. The orange points are 1cm apart and represent the
centers of the simulated taxels. The green arrows are the contact
force vectors and the red arrows are the normal components of the
contact forces.}
\vspace{-0.3cm}
\end{figure}

This testbed allows us to simulate a large number of trials. We use
the open source physics simulation library, Open Dynamics Engine
\citep{physics_ode}, to simulate a planar arm with three rotational
joints, a 1kHz joint-space impedance controller, and tactile skin
covering the entire surface of the arm with a simulated taxel
resolution of 100 taxels per meter.  \figref{fig:software_simulation}
shows a visualization of the simulated robot, tactile skin, and
taxels. 

The simulated three link planar arm has kinematics and joint limits
similar to a human operating in a plane parallel to the ground at
shoulder height with a fixed wrist. The three joints correspond to
torso rotation, shoulder, and elbow.

The arm interacts with rigid cylindrical obstacles that are either
fixed or movable. In isolation, a movable object can slide in the
plane if the force applied to it exceeds friction $(\sim2N)$, while
the fixed obstacles remain stationary regardless of the force applied
to them.

\subsection{The Robot}
\label{sec:robot}
\begin{figure}
\centering
\includegraphics[height=6.5cm]{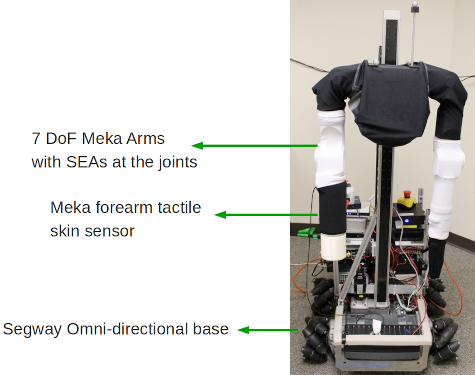}
\mycaption{
\label{fig:cody}
The robot Cody with two compliant 7 DoF arms and a tactile skin sensor
covering its right forearm.}
\end{figure}

\figref{fig:cody} shows the robot Cody that we use for experiments in
this paper. Cody has two compliant 7 DoF arms from Meka Robotics with
series elastic actuators (SEAs) for torque control at each degree of
freedom. The joint space impedance control on Cody runs at 1kHz.  Cody
has a Segway omnidirectional mobile base which we control with a PID
controller that uses visual odometry as described in our previous work
\citep{killpack2010visual}.

\begin{figure}
\centering
\includegraphics[height=3.0cm]{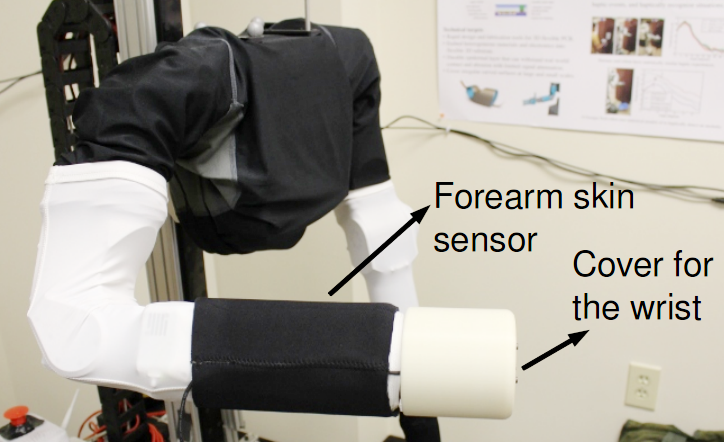}
\hspace{0.2cm}
\includegraphics[height=3.0cm]{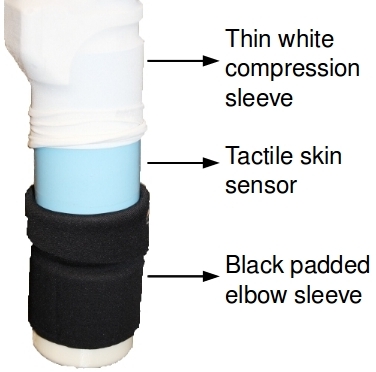}
\mycaption{
\label{fig:cody_forearm}
\textbf{Left:} Tactile skin sensor on the right forearm of Cody (underneath
the black neoprene sleeve) as well as a 3D printed cover for the
wrist. \textbf{Right:} Two additional layers (thin white compression
sleeve and black padded sleeve) that we added on top of the tactile
skin sensor (blue) once we mounted it on the robot.}
\vspace{-0.5cm}
\end{figure}

As part of this research we have developed a tactile skin sensor that
covers the right forearm of Cody, as shown in
\figref{fig:cody_forearm} and described in \secref{sec:skin}.
We currently have tactile skin covering its right forearm
only.

For experiments in realistic conditions, described in
\secref{sec:expt_real_skin}, we wanted to be able to sense contact
forces on more distal parts of the arm. To do this, we 3D printed a
cylindrical cover for the wrist of the robot, shown in
\figref{fig:cody_forearm}. We use the wrist force-torque sensor to
measure the resultant force applied to the environment by the distal
part of the arm beyond the forearm. Due to this cover, the experiments
with the forearm tactile skin sensor use only the first four degrees
of freedom of the arm.

\begin{figure*}
\centering
\includegraphics[height=4.0cm]{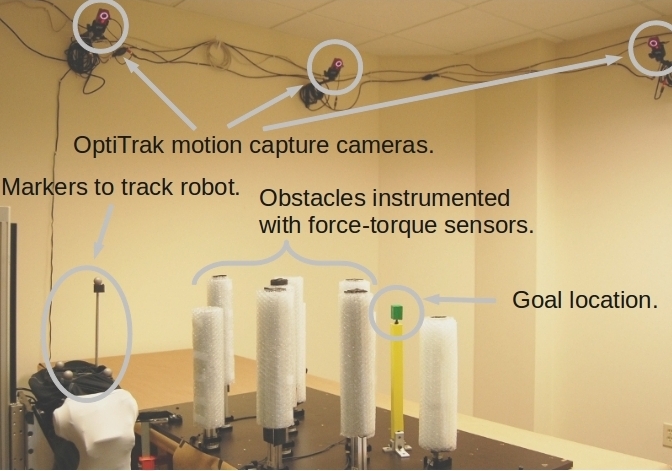}
\includegraphics[height=4.0cm]{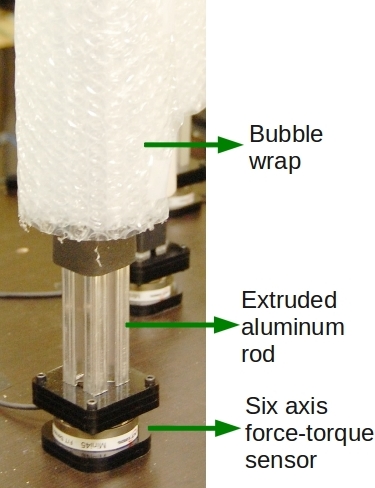}
\includegraphics[height=4.0cm]{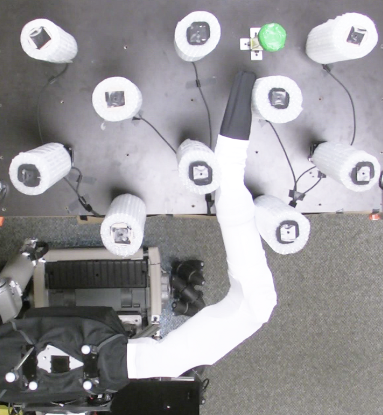}
\includegraphics[height=4.0cm]{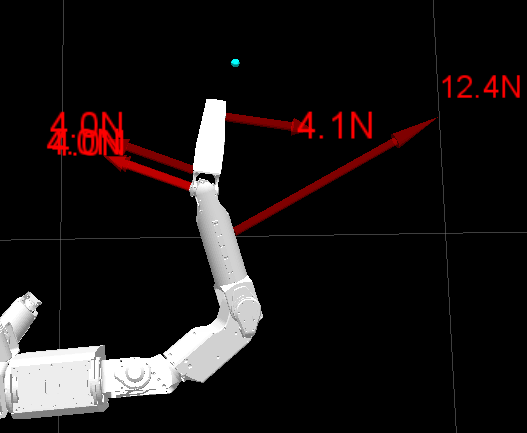}
\mycaption{\label{fig:hil_testbed} \textbf{Left:} Different
components of the hardware-in-the-loop testbed. \textbf{Middle Left:}
Close-up of one instrumented obstacle showing the force-torque sensor
at the base of an extruded aluminum rod which we have covered in
bubble wrap. \textbf{Middle Right:} Cody attempting to reach to a goal
location (green). \textbf{Right:} Visualization of the simulated
tactile skin.}
\end{figure*}

\subsection{Real Tactile Skin Sensor}
\label{sec:skin}
\figref{fig:cody_forearm} shows the tactile skin sensor that covers the
forearm of the robot Cody. Meka Robotics and the Georgia Tech
Healthcare Robotics Lab developed the forearm tactile skin sensor,
which is based on Stanford's capacitive sensing technology, as
described in \cite{ulmenskin}.

The forearm skin sensor consists of 384 taxels arranged in a $16\cross
24$ array.  There are 16 taxels along the length of the cylindrical
forearm and 24 taxels along the circumference. Each element has a
dimension of $9mm \cross 9mm$ and a sensing range of $0-30N$. We can
obtain the $16\cross 24$ taxel array sensor data at 100Hz using Robot
Operating System (ROS) drivers.


On the robot Cody, we added two additional layers on top of the
forearm skin sensor to cover the open parts of the joints, protect the
skin sensor, and make the exterior of the arm low friction. These are
shown on the right in \figref{fig:cody_forearm}. The white sleeve is a
thin neoprene McDavid compression arm sleeve, and the black layer is a
padded Ergodyne neoprene elbow sleeve designed for human athletes.

\subsection{Hardware-in-the-loop Skin Simulation}
\label{sec:hil}

Since we currently do not have whole-arm tactile skin on Cody, we have
built a hardware-in-the-loop simulation testbed to be able to simulate
whole-arm skin and test our controller on a real robot arm.

\figref{fig:hil_testbed} shows the current implementation of this
testbed. We use an OptiTrak motion tracking system to register the
positions of the obstacles, and the pose of the the robot in a common
coordinate frame.  We then use geometric collision detection from
OpenRAVE \citep{diankov2008openrave} and models of the robot arm and
obstacles to estimate the region over which each link of the robot
makes contact with the obstacles.

We have also mounted a six-axis force-torque sensor at the base of
each obstacle to measure the resultant force applied to each obstacle.
This testbed allows us to simulate skin on a real robot with 7 DoF
arms.

For every instrumented obstacle and robot link pair, we estimate at
most one contact location as the centroid of the contact region and
use the force measured by the force-torque sensor as the contact
force. If multiple links make contact with the same obstacle, we
divide the force vector's magnitude equally among all the links.

\subsection{Tactile Feedback in the Different Testbeds}
\label{sec:tactile_feedback}

There are differences in the tactile feedback in each of the three
testbeds that change the precise meaning of contact force and contact
location.

Within the software simulation testbed
(\secref{sec:software_testbed}), contact force refers to the normal
component of the force applied by the robot to the environment over
the surface covered by one taxel of the simulated tactile skin.
Contact location refers to the centroid of the simulated taxel.

Within the hardware-in-the-loop skin simulation testbed
(\secref{sec:hil}), contact force refers our estimate of the force
that a link of the real robot applies to an object. We use at most
one contact location between each link of the robot and an object
as the centroid of the contact region computed using geometric
collision detection.  We currently do not simulate individual taxels
or compute the normal component of the contact force within the
hardware-in-the-loop skin simulation testbed.

Lastly, on the real robot with the tactile skin sensor, for contacts
on the forearm of the robot, contact force refers to the normal force 
applied to the environment as measured by one taxel of the real
tactile skin sensor (\secref{sec:skin}). Contact location refers to
the centroid of the taxel. Additionally, for the distal part of the
arm beyond the forearm, we get a single resultant contact force using
a wrist force-torque sensor, as described in \secref{sec:robot}, and
we use the center of the force-torque sensor as the contact location.

\subsection{Low Stiffness at the Joints}
We use the impedance controller to maintain low stiffness at the
joints of both the real robot Cody and the robot within the software
simulation.  Within software simulation, the robot has joint stiffness
values of 30, 20, and 15 $Nm/rad$ from proximal to distal joints. These
values are similar to measured stiffnesses of humans during planar
reaching motions \citep{shadmehr1993control}.

On Cody, we use the same stiffness settings as our previous work
\citep{jain2009pon, jain2010pod}. We set the stiffness for the three
degrees of freedom in the shoulder at 20, 50, and 15 $Nm/rad$, one DoF in
the elbow at 25$Nm/rad$ and 2.5$Nm/rad$ for the wrist roll degree of
freedom. For the last two wrist joints, the robot uses position
control that relates the motor output to joint encoder readings and
ignores torque estimates from the deflection of the springs.
Consequently, the wrist is held stiff, except for the passive
compliance of the SEA springs and cables connecting the SEA to the
joints.

These stiffness settings are lower by a factor of between 400 and 1000
than the PUMA 560 manipulator \citep{kim1995configuration}. In some
postures, the stiffness at the end effector is a factor of five
lower than work on door opening with Cartesian impedance control
described in \cite{dlr_door}.

\section{Approaches used for Comparison}
\label{sec:approach_compare}
In this section we describe two approaches against which we compared
our model predictive controller's performance. The first is a baseline
controller, and the second is a state of the art geometric motion
planner that has full knowledge of the environment. We performed
comparisons against the baseline controller both on the real robot and
in software simulation. We used the geometric motion planner to
estimate optimal success rates for trials in software simulation as
detailed in \secref{sec:motion_planner}.

\subsection{Baseline Controller}
\label{sec:baseline_controller}
Our baseline controller uses the same joint-space impedance control as
the model predictive controller to maintain low stiffness at the
joints. However, it does not use feedback from the tactile skin except
to define a safety stopping criterion. Specifically, this controller
computes
\begin{align}
\Delta\phi^* &= \left( J_h^T J_h\right)^{-1} J_h^T \Delta x_d,
\label{eqn:baseline}
\end{align}
where $\Delta\phi^*\in\Re^m$ is the incremental change in the
joint-space virtual trajectory (see \figref{fig:block_diagram}), $J_h
\in\Re^{3\cross m}$ is the Jacobian at the robot's end effector (or
hand) and $\Delta x_d \in\Re^3$ is the desired Cartesian motion of the
end effector computed from \eqnref{eqn:delta_x_d}. The baseline controller
monitors the tactile skin sensor values and stops if the force at any
contact goes above the safety force threshold, $f_{c_i}^{safety}$.

If we ignore joint limits, use only $g_1$ (see
\eqnref{eqn:objective_function}) as the objective function, and the
arm is not in contact with the world, then \eqnref{eqn:baseline} is
the solution of the quadratic program for our model predictive
controller. In free-space both controllers will attempt to move the
end effector along a straight line to the goal, with identical low
stiffness settings at the joints.

%

In previous work, we have shown that a robot with low stiffness at the
joints can successfully open doors and drawers with linear virtual
trajectories for the end effector \citep{jain2009pon, jain2009bbd}.

\subsection{Motion Planner}
\label{sec:motion_planner}
For experiments in the software simulation testbed we also compare
against a bi-directional RRT motion planner as implemented in OpenRAVE
\citep{diankov2008openrave}. The motion planner has complete knowledge
of the cluttered environment and ignores movable obstacles.  We remove
the movable obstacles because the motion planner that we use does not
plan for movable obstacles.

We use the bi-directional RRT motion planner to estimate whether or
not a solution exists for a given goal location and configuration of
the clutter. We use this to estimate what the best success rate would
be for a given set of trials. It is important to note that this could
be an over estimate, since in some situations it may be impossible to
remove all of the movable obstacles, and the remaining movable
obstacles might block potential solutions.

\section{Experiments}
\label{sec:experiments}

We now describe the experiments that we performed to test our model
predictive controller.  Through these experiments, we empirically
demonstrate that our controller can effectively control three
different robot arms, a simulated 3 DoF of planar arm with simulated
tactile skin (\secref{sec:expt_software}), a real 7 DoF arm with
torque controlled joints and simulated tactile skin
(\secref{sec:expt_hil}), and the first 4 DoF of the same real robot
arm with a forearm tactile skin sensor (\secref{sec:expt_real_skin}).

We compare our controller to a motion planner (in software simulation)
and to the baseline controller within the task of reaching to a goal
location in a cluttered environment in Secs.  \ref{sec:ss_2420},
\ref{sec:hil_compare}, and \ref{sec:skin_compare}.

Additionally, we provide illustrative examples of the robot operating
in realistic conditions using the forearm tactile skin sensor
(\secref{sec:skin_realistic}), as well as ways in which the parameters
of the model predictive controller can be used to influence its
behavior, such as controlling the contact force (Secs.
\ref{sec:ss_regulate} and \ref{sec:hil_fragile}), and using online
estimates of contact stiffness to reach the goal location faster
(\secref{sec:online_stiffness}).

Due to implementation differences, the precise meaning of contact
force and contact location is different for the three testbeds, as
described in \secref{sec:tactile_feedback}.

\begin{figure*}[!t]
\centering
\includegraphics[height=4.5cm]{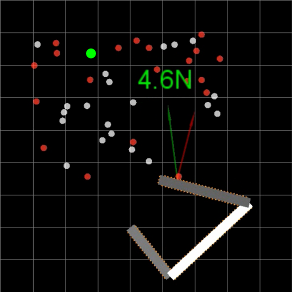}
\includegraphics[height=4.5cm]{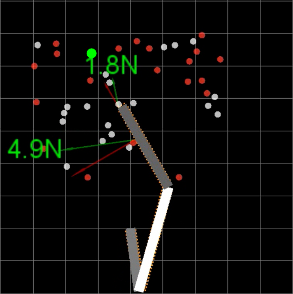}
\includegraphics[height=4.5cm]{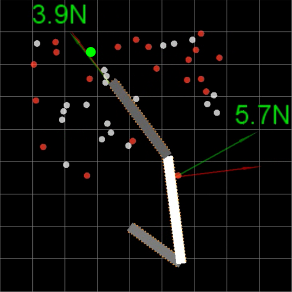}
\includegraphics[height=4.5cm]{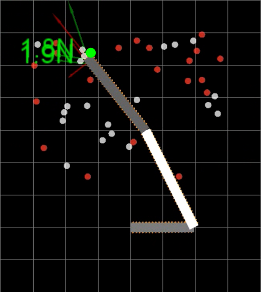}
\mycaption{\label{fig:ss_2420_montage} Sequence of images showing the
simulated robot successfully reaching to the goal location (green
circle) using MPC for one of the trials within the software simulation
testbed (see \secref{sec:ss_regulate}). The red obstacles are rigid
and fixed, while the gray obstacles are rigid and movable.}
\end{figure*}


\subsection{Pull Out and Retry}
\label{sec:pull_out_and_retry}
In the experiments described in Secs. \ref{sec:ss_2420},
\ref{sec:hil_compare}, and \ref{sec:skin_compare}, we have an
additional control layer above the model predictive controller that
makes a decision to stop the current controller if the end effector is
not moving, pulls the arm out, moves the mobile base to a different
location or selects a different starting configuration for the arm,
and retries reaching to the goal. The details of this are outside the
scope of this paper, and we treat this functionality as a black box
for the current paper.

Some of the failures in \secref{sec:ss_2420} were caused by a failure
of our current approach to pulling the arm out. Pulling out did not
fail for any of the trials described in Secs. \ref{sec:hil_compare}
and \ref{sec:skin_compare}.

\subsection{Software Simulation Testbed}
\label{sec:expt_software}
In this section we describe experiments on a large number of trials
of reaching to a goal location in an environment consisting of fixed
and movable cylindrical obstacles within the software simulation
testbed, described in \secref{sec:software_testbed} and
\figref{fig:software_simulation}. 

We generated multiple test trials by first deciding on the number of
fixed and movable obstacles. We then generated the coordinates for the
center of each cylindrical obstacle in succession by uniformly
sampling a coordinate within a fixed workspace of
0.27m$^{\textrm{2}}$. We repeated this until we found a collision free
coordinate for each obstacle in turn. We also generated a random goal
location, $x_g$, with the same sampling procedure.

If accepted for publication, we will release code, data, and
instructions to reproduce the results presented in this section.

\mysubsubsection{Comparison over 2420 Trials}
\label{sec:ss_2420}


In this experimental comparison, we selected 11 different values 
for the number of fixed and movable obstacles (from 0 to 20 in steps
of 2) and generated 20 trials for each choice of number of movable and
fixed obstacles for a total of $11\cross11\cross20 = 2420$ trials.

We compared the estimated optimal success rate using the motion
planner (\secref{sec:motion_planner}) with the baseline controller
(\secref{sec:baseline_controller}) and the model predictive
controller.

\begin{table}
\mycaption{\label{tbl:ss_controller_comparison} Results from 2420 trials in software simulation.}
\begin{center}
\begin{tabular} {|l|c|c|c|c|c|}
\hline
& Estimated & MPC (up to   & MPC & Baseline   \\  
& Optimal   & 6 Reaches) & (Single Reach)  & Controller \\  
\hline
Success rate    & 98.2\% & 91.1\% & 78.6\% & 30.5\% \\  
Avg. max. &    \multirow{2}{*}{-}   & \multirow{2}{*}{20.1N}  & \multirow{2}{*}{13.3N}
& \multirow{2}{*}{72.0N}  \\  
contact force        & & & & \\
Avg. contact & \multirow{2}{*}{-}   & \multirow{2}{*}{3.76N}  & \multirow{2}{*}{5.9N}
& \multirow{2}{*}{28.6N}  \\  
force        & & & & \\
\hline
\end{tabular}
\end{center}
\vspace{-0.3cm}
\end{table}

We allowed the model predictive controller to retry up to 5 times. If
an attempt to reach to the goal failed, the controller tried to pull
the arm out to a new starting location for the end effector, waited
for a fixed timeout period, and then retried reaching to the goal
irrespective of the success or failure of pulling out. We refer to
this as MPC with up to 6 reaches.

We set the \mytexttt{don't care force threshold}, $f_{c_i}^{thresh}$,
to $5N$ and the safety force threshold, $f_{c_i}^{safety}$ to $100N$
for each contact $c_i$ for all the trials.


Table \ref{tbl:ss_controller_comparison} shows the results from this
comparison. The model predictive controller with a single reach had a
success rate that was 48.1 percentage points more than the baseline
controller, which corresponds to a 157.7\% increase in the success
rate. For successful trials, the average speeds of the end effector
were comparable. The average speed was $0.049m/s$ for the baseline
controller and $0.043m/s$ for the model predictive controller.

Allowing the model predictive controller to retry further
increased the success rate by 12.5 percentage points (a 16\%
increase). Additionally, the estimated optimal success rate (with full
knowledge of the world and ignoring all movable obstacles) was 7.1
percentage points greater (or 7.8\% better) than the model predictive
controller with multiple reaches. Lastly, the model predictive
controller kept the contact forces lower than the baseline controller,
see Table \ref{tbl:ss_controller_comparison}.

\mysubsubsection{Regulating Contact Forces}
\label{sec:ss_regulate}

\begin{figure}[!t]
\centering
\includegraphics[height=4.2cm]{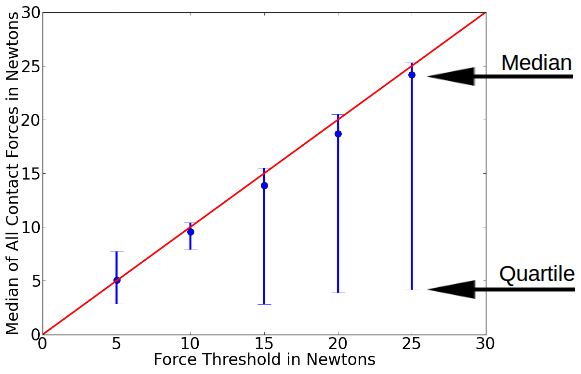}
\mycaption{
\label{fig:control_median_force}
Median, first, and fourth quartiles for the contact force for 100
trials as a function of the \mytexttt{don't care force threshold}
parameter, $f^{thresh}_{c_i}$.}
\end{figure}

As described in \secref{sec:qp_summary}, our model predictive
controller places no penalty on contact forces between zero and
$f_{c_i}^{thresh}$, the \mytexttt{don't care force threshold}.

To test the influence of $f_{c_i}^{thresh}$ experimentally, we
generated 100 trials with 20 fixed and 20 movable obstacles. We then
ran the model predictive controller with five different force
thresholds on these 100 trials, and recorded all the contact forces
every 10ms (at 100Hz). We used the same value for $f_{c_i}^{thresh}$
for each contact. \figref{fig:ss_2420_montage} shows the simulated
robot successfully reaching to the goal location for one of these
trials.

\figref{fig:control_median_force} shows the median and first and
fourth quartile of all contact force magnitudes over the 100 trials
for different values of $f_{c_i}^{thresh}$.  The correlation
coefficient between $f_{c_i}^{thresh}$ and the median force was $\geq
0.998$ providing evidence that the $f_{c_i}^{thresh}$ parameter of our
model predictive controller can be used to predictably influence the
contact forces.

\subsection{Hardware-in-the-loop Skin Simulation Testbed}
\label{sec:expt_hil}

In this section we present results using a real robot and simulated
tactile skin within the hardware-in-the-loop skin simulation testbed,
described in \secref{sec:hil} and \figref{fig:hil_testbed}.

\mysubsubsection{Online Stiffness Estimation}
\label{sec:online_stiffness}

\begin{figure}
\centering
\includegraphics[height=4.4cm]{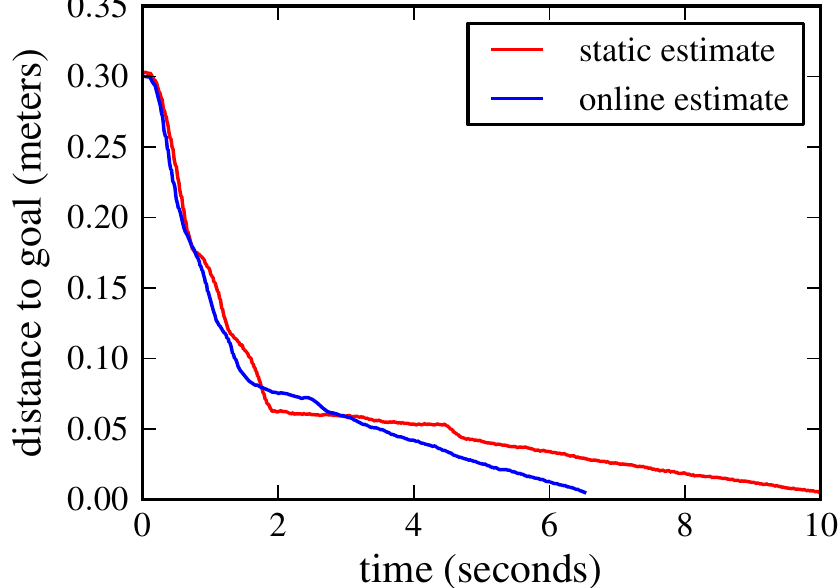}
\mycaption{
\label{fig:pillow_time_comparison}
When the robot uses online estimates of the stiffness at the contacts
(instead of conservative constants), it can push into deformable
objects more aggressively, and complete the task of reaching to a goal
location faster, shown in \figref{fig:pillow_montage}.}
\end{figure}

\begin{figure*} [!t]
\centering
\includegraphics[height=4.5cm]{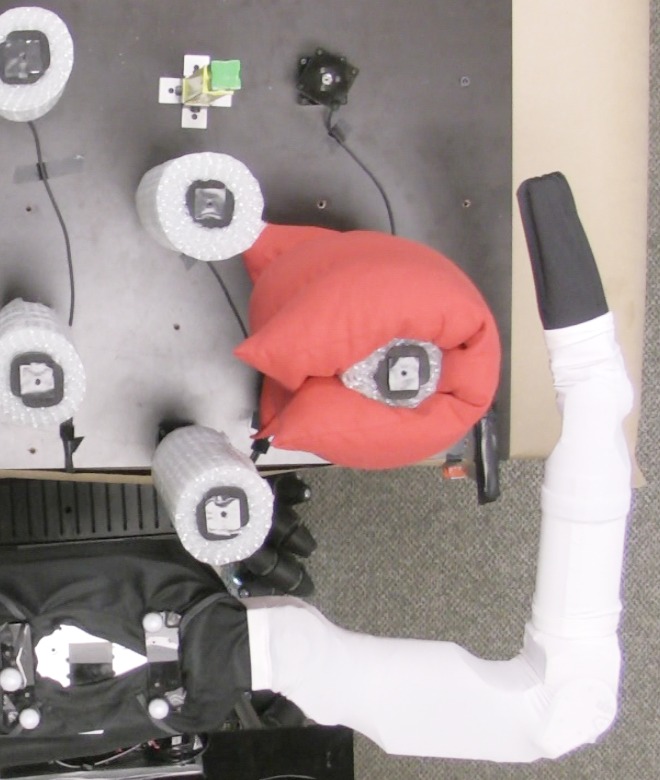}
\includegraphics[height=4.5cm]{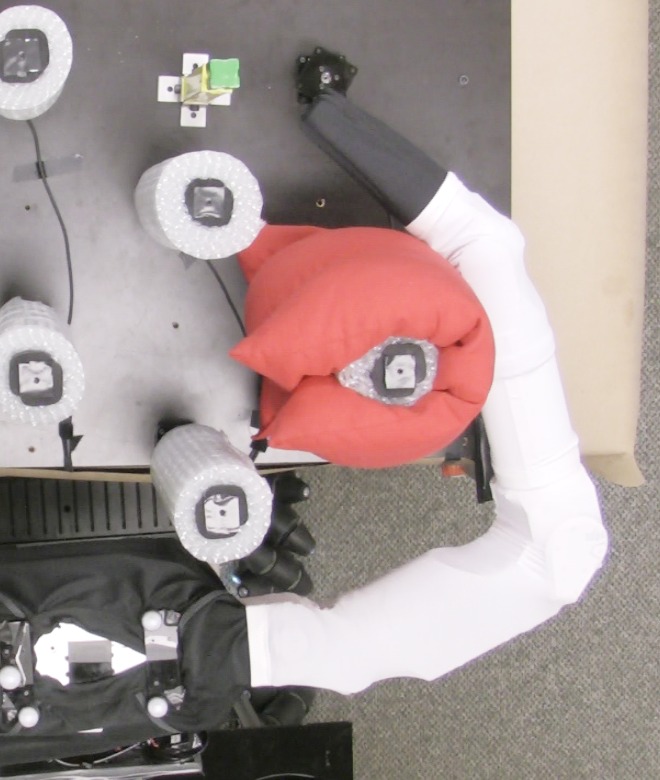}
\includegraphics[height=4.5cm]{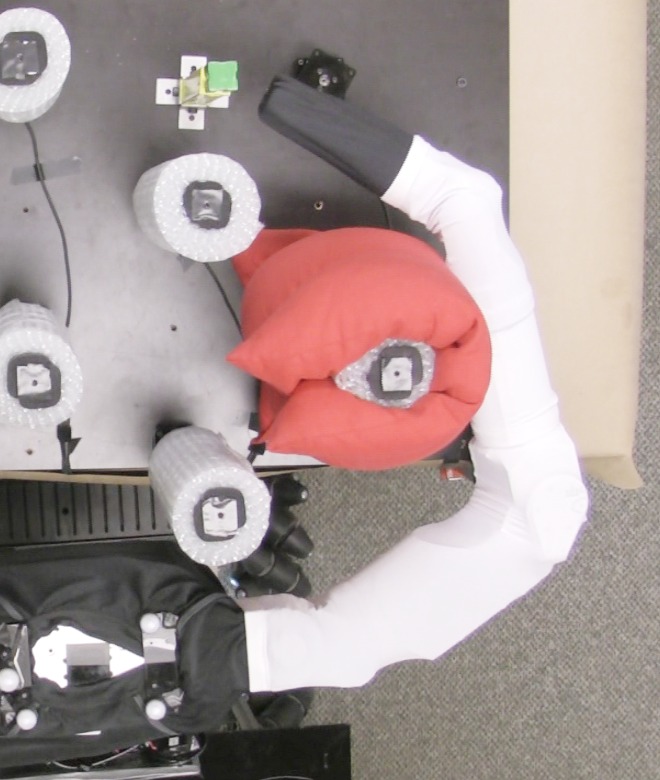}
\includegraphics[height=4.5cm]{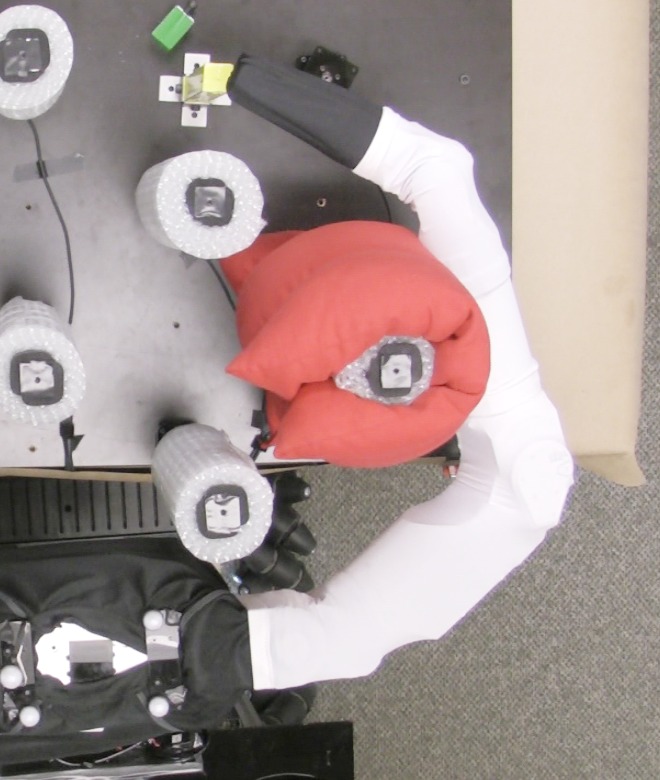}
\mycaption{\label{fig:pillow_montage} Sequence of images showing the
robot pushing on a deformable red pillow to reach to a goal
location (green).
}
\end{figure*}

One of the parameters of our model predictive controller is the
modeled stiffness at each of the contacts along the arm. This
determines how much the robot is willing to move along the contact
normal. For example, if the controller's estimate of the stiffness at
a contact is high, it will attempt to push into that contact slowly,
since its contact model will predict that a small motion will result
in a large increase in the contact force.

In this section we describe our initial efforts in estimating the
stiffness online. We set up an experiment where the robot had to push
into a deformable object (pillow) to reach to a goal location, as
shown in \figref{fig:pillow_montage}. We performed two trials. In one
trial, the robot used a static and conservative value (high stiffness)
for the stiffness at all contact locations. In the second trial, the
robot started with the same conservative estimates, but then estimated
the stiffness online while interacting with the pillow.
\figref{fig:pillow_time_comparison} shows that when the robot updated
its estimate of the stiffness, it was able to push into the pillow
more aggressively and reach the goal location faster than when the
stiffness value was a conservative static value.

To estimate the stiffness, we used a history of contact locations and
contact forces as returned by the simulated tactile skin.  We
estimated the stiffness along the current contact normal as the slope
of the line (fit using least squares) that describes the change in the
normal component of the force with motion of the contact location
along the contact normal.

A video of this experiment is part of the supplementary materials.

\begin{figure*}[!t]
\centering
\includegraphics[height=3.8cm]{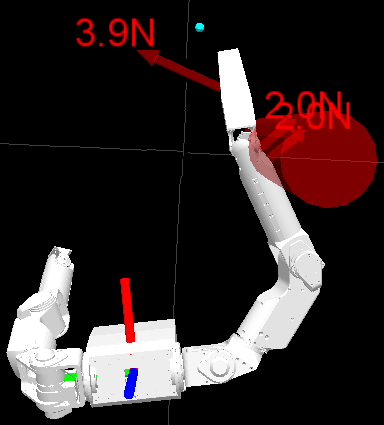}
\includegraphics[height=3.8cm]{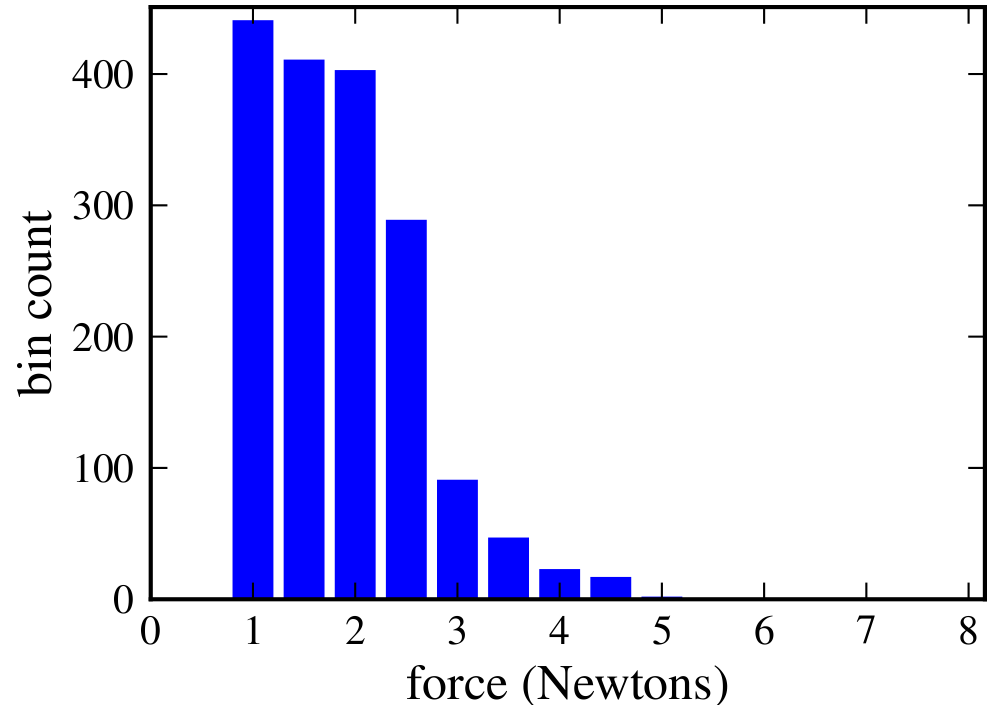}
\includegraphics[height=3.8cm]{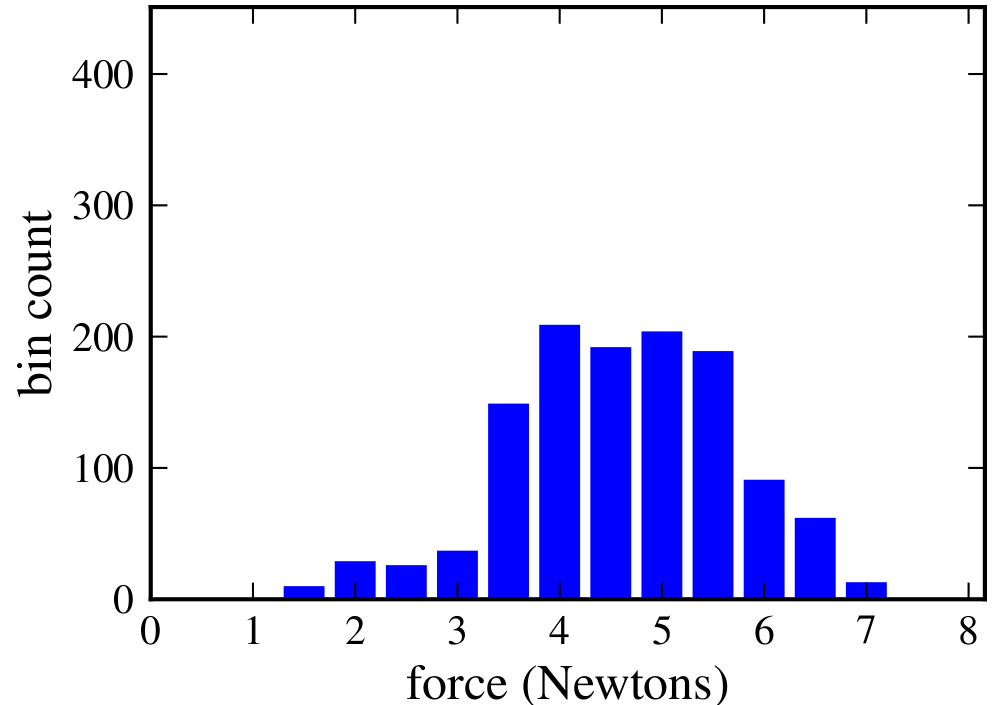}
\mycaption{
\label{fig:fragile_region_hil}
Experiment to demonstrate selective control of contact force in
different regions using the model predictive controller.
\textbf{Left:} Skin visualization when some contacts are within the
`fragile' region (red cylinder) and other contacts are outside the
`fragile' region. The cyan circle is the goal location that the robot
successfully reached. \textbf{Middle:} Histogram of contact
forces within the `fragile' region.  \textbf{Right:} Histogram of
contact forces outside the `fragile' region.}
\end{figure*}

\mysubsubsection{Selective Control of Force Applied to Different Regions
in the Environment}
\label{sec:hil_fragile}

With this experiment, we illustrate that the model predictive
controller can be used to selectively control the  contact force in
different regions. We defined a cylindrical volume in the world as
`fragile'. If the location of a contact $c_i$ in the world frame was
within the `fragile' volume, we set the \mytexttt{don't
care force threshold}, $f_{c_i}^{thresh}$, to $2N$. For contacts outside
this volume, we set it to $5N$. $f_{c_i}^{thresh}$ is used in the
inequality constraints of \eqnref{eqn:force_constraints}.

\figref{fig:fragile_region_hil} shows the forces that the robot
applied to the environment during this trial. The histograms of
contact forces within and outside the `fragile' region show that the
model predictive controller was able to selectively control the force
that it applied to different regions in the environment.

\mysubsubsection{Model Predictive Controller vs Baseline Controller}
\label{sec:hil_compare}

\begin{figure}
\centering
\includegraphics[height=3.8cm]{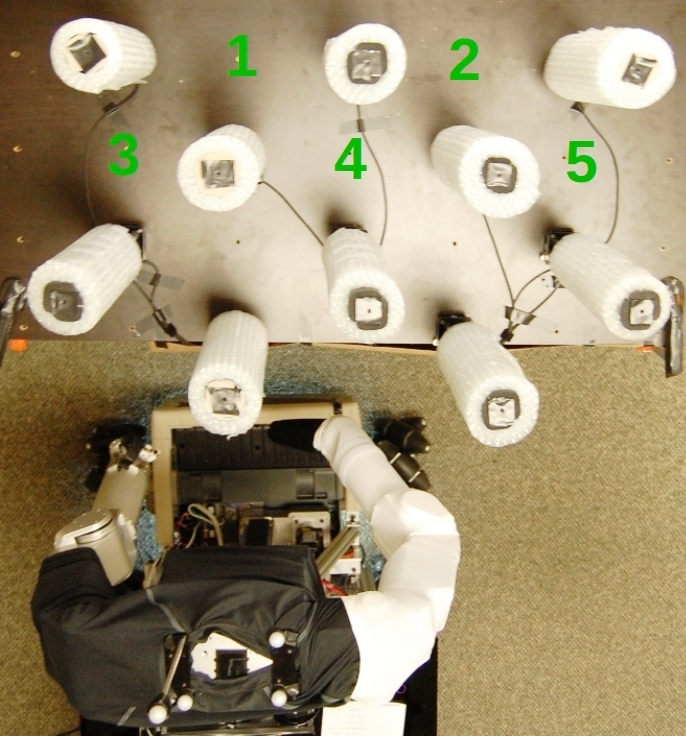}
\mycaption{
\label{fig:hil_comparison}
Five different goal locations within the hardware-in-the-loop
testbed that we used to compare the model predictive controller and
the baseline controller, described in \secref{sec:hil_compare}.}
\end{figure}

\begin{table}[t!]
\mycaption{Model predictive controller vs the
baseline controller in the hardware-in-the-loop testbed.
\label{tbl:hil_comparison}}
\begin{center}
\begin{tabular} {|l|c|c|}
\hline
& MPC &
Baseline Controller \\
\hline
Success rate & 5/5 & 3/5 \\
Avg. max. contact force & 5.6N & 17.7N \\
Avg. contact force above & \multirow{2}{*}{5.5N} &
\multirow{2}{*}{14.3N} \\
$f_{c_i}^{thresh}$ (5N) & & \\
\hline
\end{tabular}
\end{center}
\end{table}

We performed five trials with the goal location in different positions
within the hardware-in-the-loop testbed, as shown in
\figref{fig:hil_comparison}. In each trial, the robot moved its mobile
base to up to three positions equally spaced along a line and facing
the instrumented obstacles, and then attempted to reach to the goal
location from a constant pre-determined arm configuration using the
model predictive controller. The robot successfully reached each of
the five goal locations from one of the three positions. We set the
\mytexttt{don't care force threshold}, $f_{c_i}^{thresh}$, to $5N$ and
the safety force threshold $f_{c_i}^{safety}$ to $20N$ for each
contact.

As mentioned in \secref{sec:pull_out_and_retry}, we have an additional
control layer above the model predictive controller and the baseline
controller. This enables the robot to exhibit behavior such as
retrying a greedy reach from the left or right of a contact location.
A video of these five trials with MPC and the additional control layer
is part of the supplementary materials.

\begin{figure}
\centering
\includegraphics[height=2.4cm]{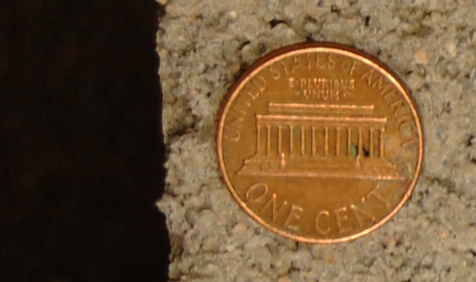}
\mycaption{
\label{fig:cinder_block_closeup}
Image showing the sharp edge and abrasive surface of the cinder
block used in the trial described in \secref{sec:skin_realistic}.}
\vspace{-0.3cm}
\end{figure}

\begin{figure*}[!t]
\centering
\includegraphics[height=5cm]{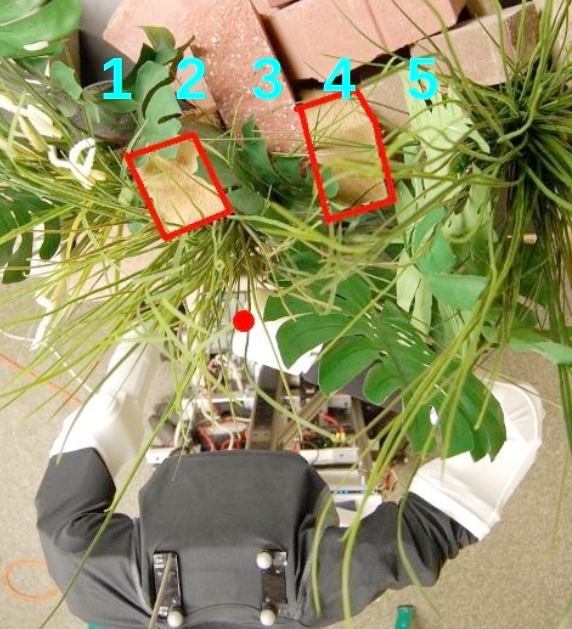}
\hspace{0.5cm}
\includegraphics[height=5cm]{figures/experiments/foliage_qp_baseline_comparison_robot_perspective}
\hspace{0.5cm}
\includegraphics[height=5cm]{figures/experiments/foliage_goal_pos_5}
\mycaption{
\label{fig:foliage_comparison}
\textbf{Left:} Five different goal locations that we used to compare
the model predictive controller and the baseline controller, described
in \secref{sec:skin_compare}. The environment consists of compliant
leaves and rigid blocks of wood (outlined in red). The red circle
denotes the position of the end effector.  \textbf{Middle:} View of
the foliage from the robot's perspective. The rigid blocks of wood are
occluded by the leaves.  \textbf{Right:} Image of the robot after it
has successfully reached goal location 5.  The red circle denotes the
end effector position.}
\end{figure*}

For each goal location, we ran the baseline controller with the
initial arm configuration and base position from which the model
predictive controller was successful. Table \ref{tbl:hil_comparison}
shows the results from the five trials. The baseline controller failed
in two out of five trials and resulted in the arm applying larger
forces on the environment. The mean contact force (including the
\mytexttt{don't care} interval of $0-5N$) was $3.3N$ for the model
predictive controller and $8.1N$ for the baseline controller.

\subsection{Forearm Tactile Skin Sensor}
\label{sec:expt_real_skin}

In this section we describe results from our experiments with the
forearm tactile skin sensor, described in \secref{sec:skin} and
\figref{fig:cody_forearm}. Since the skin sensor currently covers only
the forearm of the robot, we restricted our experiments to not have
contacts on the elbow and upper arm of the robot.

Our simulated foliage is representative of foliage found in nature. It
consists of both compliant objects (plastic leaves) and rigid and
fixed objects (blocks of wood). The leaves can result in a lot of
occlusion for conventional line of sight sensors. Furthermore, the
leaves can often be pushed aside with relatively low force but the
blocks of wood can not.

The cinder block is a rigid, heavy, and fixed object, representative
of some of the objects a robot would encounter in rubble. The diameter
of the robot's forearm ($10cm$) is close to the size of the opening of
the cinder block ($14.5cm$). Additionally, the edges are sharp and the
surface is abrasive as illustrated in
\figref{fig:cinder_block_closeup}.

\begin{figure*}[t]
\centering
\subfloat[Reaching to a goal location in foliage with multiple
contacts along the arm. The forearm and 3D printed cover for the wrist are
approximately outlined in red. \newline The goal location is vertically below
the blue bulb, and is the cyan circle in the skin
visualization.]{
\hspace{-0.8cm}
\includegraphics[height=5.5cm]{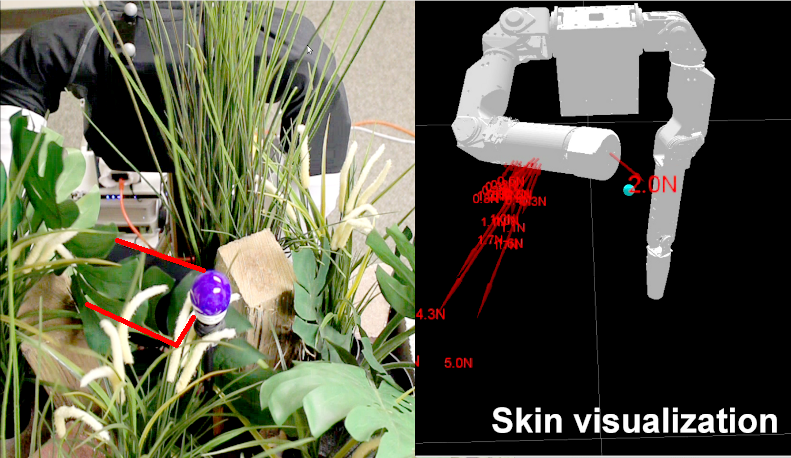}
\hspace{0.01cm}
\includegraphics[height=5.5cm]{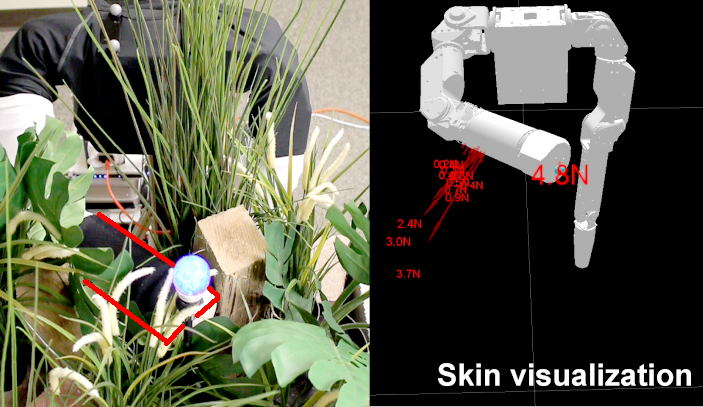}}
\\
\subfloat[Reaching to a goal location (green) through the opening of a cinder
block.]{
\hspace{-0.8cm}
\includegraphics[height=5.5cm]{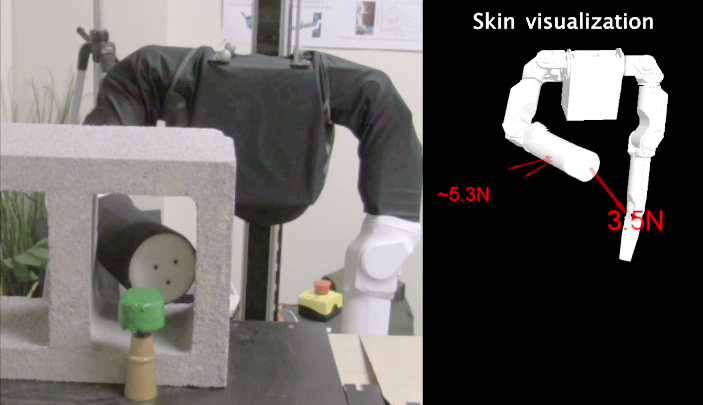}
\hspace{0.01cm}
\includegraphics[height=5.5cm]{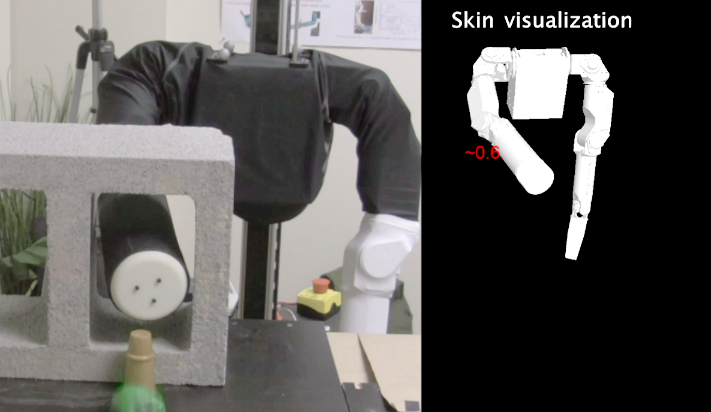}}
\\
\subfloat[Histogram of contact forces while reaching to a goal location in
foliage (left), and through the opening of the cinder block (right).]{
\includegraphics[height=4.0cm]{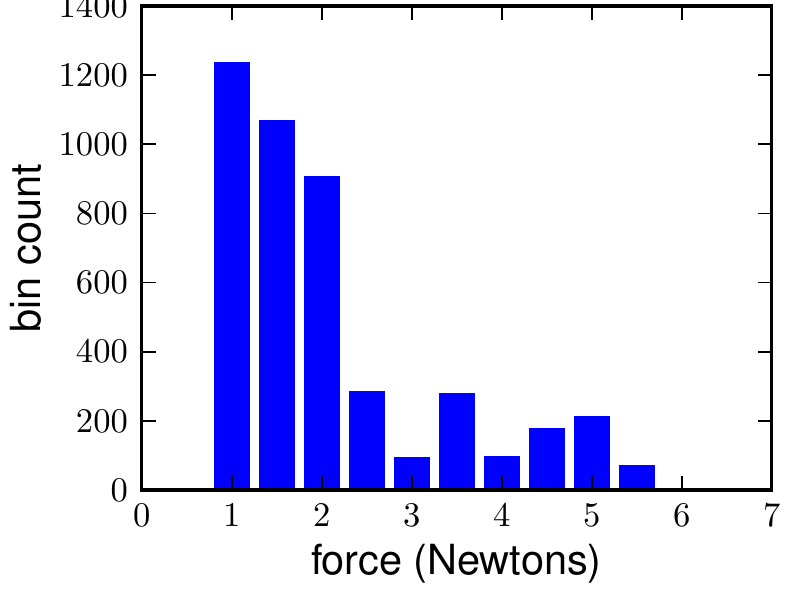}
\hspace{2.0cm}
\includegraphics[height=4.0cm]{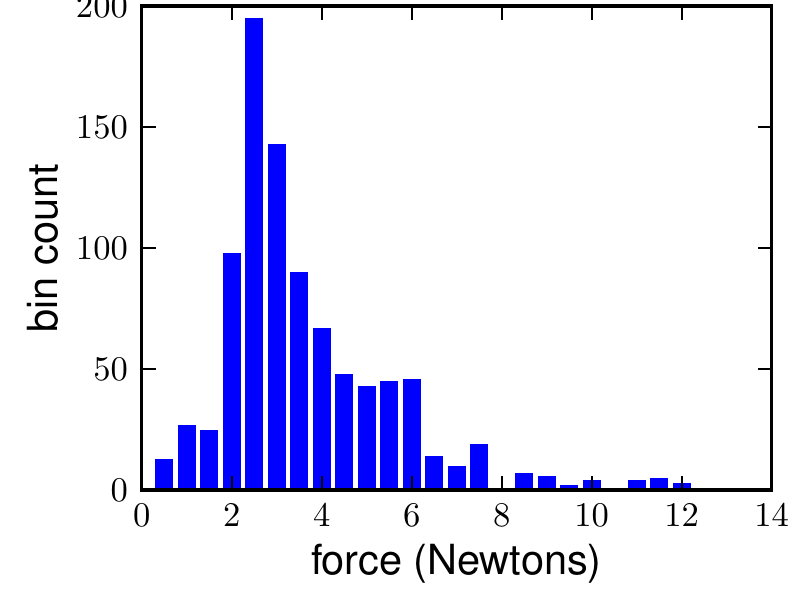}}
\mycaption{\label{fig:illustrative_cinder_block}
Cody reaching to a goal location in realistic conditions using its
forearm tactile skin sensor, described in \secref{sec:skin_realistic}.}
\vspace{-0.3cm}
\end{figure*}

\mysubsubsection{Illustrative Examples -- Foliage and Cinder Block}
\label{sec:skin_realistic}
We performed one trial each of the robot reaching to a goal location in
foliage and reaching through the opening of a cinder block.
\figref{fig:illustrative_cinder_block} shows two images and the
histograms of the contact forces for these two trials. Videos of these
two trials are part of the supplementary materials.

\mysubsubsection{Model Predictive Controller vs Baseline Controller in
Foliage}
\label{sec:skin_compare}
For a more careful evaluation of the model predictive controller and
the baseline controller in realistic conditions using the forearm
tactile skin sensor, we performed five trials with automatically
generated goal locations that were equally spaced along a line within
our simulated foliage, as shown in \figref{fig:foliage_comparison}.

We started each trial by positioning the robot at the same location in
front of the clutter. The robot then autonomously moved its mobile
base to four roughly equally spaced positions along a line, and
attempted to reach to the goal location using both the model
predictive controller and the baseline controller, as described in
\figref{fig:foliage_5_trial_block}.

\begin{figure}
\small
\centering
\begin{tikzpicture}[auto, node distance=0.8cm,>=latex']
    \node [input, name=input] {};
    \node [clearblock, below of=input, node distance=0.6cm] (move_base) {Move to next base position};
    \node [clearblock, below of=move_base] (reach_qp) {Reach to goal
        with MPC};
    \node [clearblock, below of=reach_qp] (out1) {Pull out the arm};
    \node [clearblock, below of=out1] (reach_baseline) {Reach to goal with
        Baseline controller};
    \node [clearblock, below of=reach_baseline] (out2) {Pull out the arm};
    \node [output, below of=out2, name=output, node distance=0.4cm] {};
    \node [output, left of=reach_qp, name=dummy, node distance=3.0cm] {};

    \draw [myarrow] (input) -- (move_base);
    \draw [myarrow] (move_base) -- (reach_qp);
    \draw [myarrow] (reach_qp) -- (out1);
    \draw [myarrow] (out1) -- (reach_baseline);
    \draw [myarrow] (reach_baseline) -- (out2);
    \draw [myline] (out2) -- (output);
    \draw [myline] (output) -| (dummy);
    \draw [myline] (dummy) |- (input);
\end{tikzpicture}
\mycaption{
\label{fig:foliage_5_trial_block}
Different steps that the robot performed for each of the five trials
in the model predictive controller vs baseline controller comparison
in foliage, described in \secref{sec:skin_compare}}
\end{figure}

\begin{table} [t]
\mycaption{Model predictive controller vs baseline controller
in foliage.\label{tbl:foliage_comparison}}
\vspace{-0.2cm}
\begin{center}
\begin{tabular} {|l|c|c|}
\hline
& MPC &
Baseline Controller \\
\hline
Success rate & 3/5 & 1/5 \\
Exceeded safety threshold (15N) & 0/20 attempts & 19/20 attempts \\
Avg. max. contact force & 5.5N & 14.5N \\
Avg. contact force above & \multirow{2}{*}{5.2N} & \multirow{2}{*}{9.2N} \\
$f_{c_i}^{thresh}$ (5N) & & \\
\hline
\end{tabular}
\end{center}
\vspace{-0.3cm}
\end{table}

Table \ref{tbl:foliage_comparison} shows the results from a total of
twenty reach attempts for each controller from the five trials.  The
model predictive controller successfully reached goal locations 1, 3,
and 5, while the baseline controller was only successful for goal
location 5. Further, the model predictive controller successfully kept
the contact forces around the \mytexttt{don't care force threshold},
$f_{c_i}^{thresh}$, of $5N$. In contrast, the baseline controller
exceeded the safety force threshold, $f_{c_i}^{safety}$, of $15N$, 19
out of 20 times. The mean contact force (including the \mytexttt{don't
care} interval of $0-5N$) was $3.2N$ for the model predictive
controller and $4.5N$ for the baseline controller.

A video of these five trials is part of the supplementary materials.

\section{Discussion}
\label{sec:discussion}

Within this section, we discuss broader implications of our research,
future work, and current limitations.

\subsection{The Big Picture}

Our results suggest that our approach is well-matched to manipulation
in real-world, high-clutter environments, although further evaluation
is required. There are also several broader implications of our work,
which we now discuss.

\mysubsubsection{Greedy Control}

The performance of our system suggests that greedy feedback control
can perform well in practice and that detailed models of the
environment and long time horizon planning may not be necessary to
achieve high performance. This is similar in spirit to some research
on bipedal walking described in \cite{byl2008approximate}.

From our perspective, if empirical research
with real robots continues to support this conjecture, it would be a
welcome outcome. If one considers the complexity associated with
natural environments, such as swamps, rainforests, and caves, a
requirement for detailed models and long time horizon planning seems
extremely daunting, if not infeasible. Fluids, gasses, granular media,
biological materials, and active agents are just a few of the complex
contents found across the earth. That biological organisms of all
shapes and sizes regularly perform impressive feats of manipulation in
these environments demonstrates that the problems are not intractable
and that biology has found solutions worthy of emulation. Compliant
actuation and whole-body tactile sensing combined with a willingness
to make contact with the unknown may be important characteristics of
the biological solution to manipulation.

\mysubsubsection{Reaching into the Unknown}

Our results also suggest that reaching into the unknown can be a
reasonable action for robots with compliant joints and whole-body
tactile sensing. As more robots with these underlying capabilities
emerge, the value of these attributes should become clearer,
especially given the current rarity of whole-body tactile sensing. So
far, we have demonstrated the feasibility of haptically reaching into
grass-like vegetation with hidden wooden objects, into a constrained
massive cinder block with coarse and sharp edges, and into a field of
rigid posts covered with compliant materials. In these experiments,
both the robot and the environment were unscathed in spite of repeated
reaches without explicit foreknowledge of the environments'
contents. Demonstrating success with more diverse environments, real
natural environments, real tactile sensing across the entire robot
arm, and higher usable degrees of freedom will be important future
work.

\mysubsubsection{Human Environments}

Within this paper, we have frequently referred to natural high-clutter
outdoor environments, such as foliage, in part because of our
biological inspiration. However, we expect that our approach and
methods would also be beneficial to manipulation in everyday human
environments. Humans often encounter high clutter, such as collections
of objects on top of tables and shelves, and inside drawers and other
containers. Humans also reach into constrained volumes, such as when
retrieving objects from under furniture, cleaning hard to reach areas,
or performing maintenance on machinery. We would expect service robots
to benefit from comparable capabilities. Assistive robots might also
benefit from our approach, since humans often make contact with their
arms and other parts of their body when providing physical assistance
to other people, such as when helping someone get out of bed. 


\mysubsubsection{Emergent Intelligence}

Although our low-level controller is greedy and has been provided
waypoints that are always along a straight line from the end effector's
current position to the goal location, to us the resulting qualitative
motion of the robot's arm appears to be intelligent, complex, and
lifelike. To objectively support these notions would likely require
human-robot interaction studies, so they must be treated
skeptically. Nonetheless, like Herbert Simon's ant walking on the
beach \citep{simon1996sciences}, the robot's reactions to the
complexity of the world result in complex emergent motion. For
example, due to tactile sensing and the controller, the robot can move
against a compliant object until the force is higher than desired and
then pivot around it. Maneuvers such as this appear to be sensible,
even though they are not the result of explicitly planned
trajectories. Likewise, the robot can easily respond to dynamic
elements of the environment, since it regenerates a model at each time
step based on its tactile sensing and greedily decides how to move.
Our approach and results relate strongly to behavior-based robotics
\citep{brooks_intelligence_without_reason}.

\subsection{Future Work}

The controller we have presented has promising performance and its
properties serve to illustrate our overall approach. Many
opportunities exist to integrate this controller, or similar
controllers, into manipulation systems. We have presented results with
reactive behaviors, but we would expect the controller to also be
appropriate for the execution of planned trajectories or commands from
a teleoperator. Due to the greedy controller and the potential for
local minima, some form of higher level control is required. Our
approach has been to develop higher level controllers that detect when
the arm has stopped making progress (reached a local minimum), and
then restart the controller with new initial conditions. How to best
design complementary higher level controllers and associated
representations with memory merits further inquiry.

There are also numerous avenues that remain open for further
development and evaluation of this controller and similar
controllers. For example, we have fixed the stiffness of the robot's
joints to low constant values, which could instead be varied at each
time step. A related open question is how to initialize and adapt the
various controller parameters given a robot, an environment, and a
task. Data-driven methods from machine learning might be a worthwhile
direction for research related to this question.

\subsection{Limitations}

In spite of its good performance in our experiments, the current
controller does have limitations that could motivate revisions of this
controller, or new controllers entirely.  First, our contact model
consists of a linear spring, which is computationally favorable, but
predicts adhesive forces when breaking contact. Second, the controller
places no penalty on a predicted contact force that is less than
$f_{c_i}^{thresh}$, and has a hard inequality constraint that prevents
higher predicted forces. Yet in practice, the actual contact forces
sometimes exceed this constraint. Currently, the controller handles
this by removing the constraint and adding a quadratic penalty, which
results in an objective function that varies over time. It may be
advantageous to instead use a constant objective function that is
smooth, hence softening the constraint.

Third, the current controller ignores dynamics. The resulting
quasi-static model is well-matched to slow motions. And, slow motions
are reasonable when performing haptically-guided manipulation without
a model in high clutter, since a collision could occur at any
moment. Nonetheless, taking dynamics into account might enable the
controller to attain better performance at higher speeds, and better
control of the arm's velocity. Fourth, so far, we have only tested the
controller for achieving a position of the end effector. Objectives
such as an arm posture or full pose of the end effector would be
better matched to some tasks. These and other objectives could
plausibly be represented as quadratic objective functions, but we have
not tested this possibility.



\section{Conclusion}
\label{sec:conclusion}

We have presented our approach to manipulation, which from the outset
emphasizes contact with the world. We assume that low contact forces
are benign, and focus on the development of robotic systems that can
control their contact forces during goal-directed motion. Inspired by
biology, we assume that the robot has low-stiffness compliant
actuation at its joints, and tactile sensing across its entire
surface.

We then described a novel controller that exploits these
assumptions. The controller only requires haptic sensing and does not
need a detailed model of the environment prior to contact. It also
explicitly allows multiple contacts across the entire surface of the
arm. 

The controller uses model predictive control (MPC) with a time horizon
of length one, and a linear quasi-static model. As quantitatively
summarized in the following list, we have empirically shown that our
MPC controller enables a variety of robots to haptically reach goal
locations in highly cluttered environments with low contact forces,
and that it outperforms a baseline controller that uses the same
low-stiffness actuation at its joints.

\begin{enumerate}

\item \textbf{Successfully enables a simulated robot to reach to a
  location in clutter:} In an experiment with 2420 trials, our MPC
  controller succeeded in 150\% more trials than our baseline
  controller, and had lower average contact forces ($5.9N$ vs.
  $28.6N$). Even though it is a greedy controller, it was within
  $\sim$8\% of optimal performance when allowed to make up to 6 reach
  attempts. In another experiment, the correlation between the MPC
  controller's \mytexttt{don't care force threshold} and the median
  applied force was $\geq$0.998, which indicates that this controller
  parameter predictably influences the contact forces applied by the
  arm.

\item \textbf{Successfully enables a real robot with simulated skin to
  reach to a location in clutter:} Using the MPC controller, our robot
  autonomously reached 5 human specified targets, while the baseline
  controller only reached 3. The MPC controller had lower average
  maximum force for reach attempts ($5.6N$ vs. $17.7N$ for baseline).
  In addition, we demonstrated that the robot can apply less force to
  a designated fragile region, and can estimate that a contact has low
  stiffness online resulting in more aggressive and efficient progress
  to a goal location.

\item \textbf{Successfully enables a real robot with real skin to
  reach to a location in real clutter:} We performed a
  \textit{fully autonomous evaluation} of our MPC controller
  in which it successfully commanded the real robot with real forearm
  skin to reach 3 out of 5 \textit{automatically generated
  target locations} within foliage (target locations were not
  necessarily achievable and could be embedded within rigid
  objects). The baseline controller succeeded in reaching 1 out of 5
  of these targets from the same starting conditions as the MPC
  controller. The MPC controller also achieved lower average maximum
  force than the baseline controller ($5.5N$ vs.  $14.5N$), which
  corresponds to the MPC controller's \textit{don't care force
  threshold} of $5N$. 

  The MPC controller also enabled the robot to reach into a cinder
  block representative of rubble, which demonstrates the feasibility
  of moving against rigid, sharp and coarse materials, and through
  constrained passages.

\end{enumerate}


\section{Supplementary Materials}
\mysubsubsection{Videos}
We have prepared the following videos as part of the supplementary
materials:
\begin{itemize}
\item Model predictive controller vs baseline controller
within the hardware-in-the-loop testbed as described in
\secref{sec:hil_compare}.
\item Online stiffness estimation with the hardware-in-the-loop
testbed as described in \secref{sec:online_stiffness}.
\item Illustrative examples of reaching in foliage and through the
opening of a cinder block using the real forearm tactile skin sensor,
described in \secref{sec:skin_realistic}.
\item Model predictive controller vs baseline controller in foliage
using the forearm tactile skin sensor, as described in
\secref{sec:skin_compare}.
\item Video showing the ``simple'' impedance controller and low
stiffness at the joints for the robot Cody.
\item The simulated robot reaching to the goal location in one of the
trials with the software simulation testbed described in
\secref{sec:ss_regulate}.
\end{itemize}

\mysubsubsection{Code} If accepted for publication, we will release
our code as open source. We will also provide instructions and data to
reproduce the results within the software simulation testbed
(\secref{sec:expt_software}).

\section{Acknowledgements}
We gratefully acknowledge support from DARPA Maximum Mobility and
Manipulation (M3) Contract W911NF-11-1-603. This work benefitted from
discussions with Magnus B. Egerstedt, Harvey Lipkin, Mike Stilman, and
James M. Rehg.

We thank Mark Cutkosky and the Stanford Biomimetics and Dexterous
Manipulation Lab for their contributions to the forearm tactile skin
sensor.

We also thank Jeff Weber, Andy Metzger, Benjamin Valenti, Pierre-Luc
Bacon, Robert Kelbley, and John Ulmen for their contributions to the
forearm tactile skin sensor hardware and software.


\bibliographystyle{spbasic}
\bibliography{qp}

\begin{thebibliography}{103}
\providecommand{\natexlab}[1]{#1}
\providecommand{\url}[1]{{#1}}
\providecommand{\urlprefix}{URL }
\expandafter\ifx\csname urlstyle\endcsname\relax
  \providecommand{\doi}[1]{DOI~\discretionary{}{}{}#1}\else
  \providecommand{\doi}{DOI~\discretionary{}{}{}\begingroup
  \urlstyle{rm}\Url}\fi
\providecommand{\eprint}[2][]{\url{#2}}

\bibitem[{Abbeel et~al(2010)Abbeel, Coates, and Ng}]{abbeel2010autonomous}
Abbeel P, Coates A, Ng A (2010) Autonomous helicopter aerobatics through
  apprenticeship learning. The International Journal of Robotics Research

\bibitem[{Albu-Schaffer et~al(2003)Albu-Schaffer, Ott, Frese, and
  Hirzinger}]{albu2003cartesian}
Albu-Schaffer A, Ott C, Frese U, Hirzinger G (2003) Cartesian impedance control
  of redundant robots: Recent results with the dlr-light-weight-arms. In:
  Robotics and Automation, 2003. Proceedings. ICRA'03. IEEE International
  Conference on, IEEE, vol~3, pp 3704--3709

\bibitem[{Alexander(1990)}]{alexander1990three}
Alexander R (1990) Three uses for springs in legged locomotion. The
  International Journal of Robotics Research

\bibitem[{Bellingham et~al(2002)Bellingham, Richards, and
  How}]{bellingham2002receding}
Bellingham J, Richards A, How J (2002) Receding horizon control of autonomous
  aerial vehicles. In: American Control Conference

\bibitem[{Bianchi(2007)}]{mechanotransduction2007}
Bianchi L (2007) Mechanotransduction: Touch and feel at the molecular level as
  modeled in caenorhabditis elegans. Molecular Neurobiology 36(3):254--271

\bibitem[{Bicchi(1993)}]{bicchi1993force}
Bicchi A (1993) Force distribution in multiple whole-limb manipulation. In:
  IEEE International Conference on Robotics and Automation

\bibitem[{Bicchi and Kumar(2000)}]{bicchi2000robotic}
Bicchi A, Kumar V (2000) Robotic grasping and contact: A review. In: IEEE
  International Conference on Robotics and Automation.

\bibitem[{Bicchi et~al(1993)Bicchi, Salisbury, and Brock}]{bicchi1993contact}
Bicchi A, Salisbury J, Brock D (1993) Contact sensing from force measurements.
  The International Journal of Robotics Research

\bibitem[{Boyd and Vandenberghe(2004)}]{boyd2004convex}
Boyd S, Vandenberghe L (2004) Convex optimization. Cambridge Univ Pr

\bibitem[{Brooks(1991)}]{brooks_intelligence_without_reason}
Brooks R (1991) Intelligence without reason. Artificial intelligence: critical
  concepts

\bibitem[{Buerger(2006)}]{buerger2006stable}
Buerger S (2006) {Stable, High-Force, Low-Impedance Robotic Actuators for
  Human-Interactive Machines}. PhD thesis, MIT

\bibitem[{Buerger and Hogan(2007)}]{buerger2007complementary}
Buerger S, Hogan N (2007) Complementary stability and loop shaping for improved
  human--robot interaction. IEEE Transactions on Robotics 23(2):232--244

\bibitem[{Byl and Tedrake(2008)}]{byl2008approximate}
Byl K, Tedrake R (2008) Approximate optimal control of the compass gait on
  rough terrain. In: IEEE International Conference on Robotics and Automation

\bibitem[{Catania(1999)}]{catania1999nose}
Catania K (1999) A nose that looks like a hand and acts like an eye: the
  unusual mechanosensory system of the star-nosed mole. Journal of Comparative
  Physiology A: Neuroethology, Sensory, Neural, and Behavioral Physiology

\bibitem[{De~Schutter et~al(1999)De~Schutter, Bruyninckx, Dutr{\'e}, De~Geeter,
  Katupitiya, Demey, and Lefebvre}]{de1999estimating}
De~Schutter J, Bruyninckx H, Dutr{\'e} S, De~Geeter J, Katupitiya J, Demey S,
  Lefebvre T (1999) Estimating first-order geometric parameters and monitoring
  contact transitions during force-controlled compliant motion. The
  International Journal of Robotics Research

\bibitem[{Diankov and Kuffner(2008)}]{diankov2008openrave}
Diankov R, Kuffner J (2008) Openrave: A planning architecture for autonomous
  robotics. Robotics Institute, Pittsburgh, PA, Tech Rep CMU-RI-TR-08-34

\bibitem[{Dogar and Srinivasa(2011)}]{dogar2011fpc}
Dogar M, Srinivasa S (2011) A framework for push-grasping in clutter. Robotics:
  Science and Systems

\bibitem[{Dogar et~al(2010)Dogar, Hemrajani, Leeds, Kane, and
  Srinivasa}]{dogar2010}
Dogar M, Hemrajani V, Leeds D, Kane B, Srinivasa S (2010) {Proprioceptive
  Localization for Mobile Manipulators}. Tech. rep., Carnegie Mellon University

\bibitem[{Dominy(2004)}]{dominy2004fruits}
Dominy N (2004) Fruits, fingers, and fermentation: the sensory cues available
  to foraging primates. Integrative and Comparative Biology

\bibitem[{Eberman(1989)}]{eberman1989whole}
Eberman B (1989) {Whole-arm manipulation: kinematics and control}. Master's
  thesis, MIT

\bibitem[{Eberman and Salisbury(1990)}]{eberman1990determination}
Eberman B, Salisbury J (1990) Determination of manipulator contact information
  from joint torque measurements. In: Experimental Robotics I

\bibitem[{Edsinger and Kemp(2007{\natexlab{a}})}]{kemp_roman_2007}
Edsinger A, Kemp CC (2007{\natexlab{a}}) Human-robot interaction for
  cooperative manipulation: Handing objects to one another. In: Proceedings of
  the 16th IEEE International Symposium on Robot and Human Interactive
  Communication (RO-MAN)

\bibitem[{Edsinger and Kemp(2007{\natexlab{b}})}]{kemp_icar_2007}
Edsinger A, Kemp CC (2007{\natexlab{b}}) Two arms are better than one: A
  behavior-based control system for assistive bimanual manipulation. In:
  Proceedings of the 13th International Conference on Advanced Robotics (ICAR)

\bibitem[{Erez et~al(2011)Erez, Tassa, and Todorov}]{erezinfinite}
Erez T, Tassa Y, Todorov E (2011) Infinite-horizon model predictive control for
  periodic tasks with contacts. In: Robotics: Science and Systems (RSS)

\bibitem[{Escande and Kheddar(2009)}]{escande2009contact}
Escande A, Kheddar A (2009) Contact planning for acyclic motion with tasks
  constraints. In: IEEE/RSJ International Conference on Intelligent Robots and
  Systems (IROS)

\bibitem[{Featherstone and Orin(2008)}]{dynamics_springer_handbook}
Featherstone R, Orin DE (2008) Chapter 2: Dynamics, Handbook of Robotics,
  Siciliano, Bruno; Khatib, Oussama (Eds.). Springer

\bibitem[{Frank et~al(2011)Frank, Stachniss, Abdo, and
  Burgard}]{frank2011using}
Frank B, Stachniss C, Abdo N, Burgard W (2011) Using gaussian process
  regression for efficient motion planning in environments with deformable
  objects. In: Workshops at the Twenty-Fifth AAAI Conference on Artificial
  Intelligence

\bibitem[{From et~al(2011)From, Gravdahl, Lillehagen, and
  Abbeel}]{from2011motion}
From P, Gravdahl J, Lillehagen T, Abbeel P (2011) Motion planning and control
  of robotic manipulators on seaborne platforms. Control engineering practice

\bibitem[{Garcia et~al(1989)Garcia, Prett, and Morari}]{garcia1989model}
Garcia C, Prett D, Morari M (1989) Model predictive control: Theory and
  practice--a survey. Automatica

\bibitem[{Garcia et~al(1998)Garcia, Chatterjee, Ruina, and
  Coleman}]{garcia1998simplest}
Garcia M, Chatterjee A, Ruina A, Coleman M (1998) The simplest walking model:
  Stability, complexity, and scaling. Journal of Biomechanical Engineering

\bibitem[{Goodman(2006)}]{wormbook2006}
Goodman MB (2006) WormBook, chap Mechanosensation.
  \doi{10.1895/wormbook.1.7.1}, \urlprefix\url{http://www.wormbook.org/}

\bibitem[{Gu and Ballard(2006)}]{gu2006epb}
Gu X, Ballard D (2006) {An equilibrium point based model unifying movement
  control in humanoids}. In: RSS

\bibitem[{Hauser et~al(2005)Hauser, Bretl, and Latombe}]{hauser2005non}
Hauser K, Bretl T, Latombe J (2005) {Non-gaited humanoid locomotion planning}.
  In: Humanoids

\bibitem[{Hersch and Billard(2006)}]{hersch2006biologically}
Hersch M, Billard A (2006) A biologically-inspired controller for reaching
  movements. In: IEEE/RAS-EMBS International Conference on Biomedical Robotics
  and Biomechatronics (BIOROB)

\bibitem[{Hogan(1984)}]{hogan1984adaptive}
Hogan N (1984) Adaptive control of mechanical impedance by coactivation of
  antagonist muscles. IEEE Transactions on Automatic Control

\bibitem[{Hogan(1988)}]{hogan1988stability}
Hogan N (1988) On the stability of manipulators performing contact tasks.
  Robotics and Automation, IEEE Journal of 4(6):677--686

\bibitem[{Hogan and Buerger(2005)}]{hogan2005impedance}
Hogan N, Buerger S (2005) {Impedance and Interaction Control}, Robotics and
  Automation Handbook, chap~19

\bibitem[{Hsiao and Lozano-Perez(2006)}]{hsiao2006imitation}
Hsiao K, Lozano-Perez T (2006) Imitation learning of whole-body grasps. In:
  IEEE/RSJ International Conference on Intelligent Robots and Systems

\bibitem[{Hsiao et~al(2010)Hsiao, Chitta, Ciocarlie, and
  Jones}]{hsiao2010contact}
Hsiao K, Chitta S, Ciocarlie M, Jones E (2010) Contact-reactive grasping of
  objects with partial shape information. In: IEEE/RSJ International Conference
  on Intelligent Robots and Systems (IROS)

\bibitem[{Ivaldi et~al(2010)Ivaldi, Fumagalli, Nori, Baglietto, Metta, and
  Sandini}]{ivaldi2010approximate}
Ivaldi S, Fumagalli M, Nori F, Baglietto M, Metta G, Sandini G (2010)
  Approximate optimal control for reaching and trajectory planning in a
  humanoid robot. In: IEEE/RSJ International Conference on Intelligent Robots
  and Systems (IROS)

\bibitem[{Iwaniuk and Whishaw(1999)}]{iwaniuk1999skilled}
Iwaniuk A, Whishaw I (1999) How skilled are the skilled limb movements of the
  raccoon (procyon lotor)? Behavioural brain research 99(1):35--44

\bibitem[{J.~Pratt and Pratt(2001)}]{pratt_vmc}
J~Pratt PD M~Chew, Pratt G (2001) {Virtual Model Control}: An intuitive
  approach for bipedal locomotion. International Journal of Robotics Research
  20(2):129--143

\bibitem[{Jain and Kemp(2009{\natexlab{a}})}]{jain2009bbd}
Jain A, Kemp CC (2009{\natexlab{a}}) {Behavior-based door opening with
  equilibrium point control}. In: RSS Workshop: Mobile Manipulation in Human
  Environments

\bibitem[{Jain and Kemp(2009{\natexlab{b}})}]{jain2009pon}
Jain A, Kemp CC (2009{\natexlab{b}}) {Pulling Open Novel Doors and Drawers with
  Equilibrium Point Control}. In: Humanoids

\bibitem[{Jain and Kemp(2010{\natexlab{a}})}]{jain2010auro}
Jain A, Kemp CC (2010{\natexlab{a}}) {EL-E: An Assistive Mobile Manipulator
  that Autonomously Fetches Objects from Flat Surfaces}. Autonomous Robots

\bibitem[{Jain and Kemp(2010{\natexlab{b}})}]{jain2010pod}
Jain A, Kemp CC (2010{\natexlab{b}}) {Pulling Open Doors and Drawers:
  Coordinating an Omni-directional Base and a Compliant Arm with Equilibrium
  Point Control}. In: ICRA

\bibitem[{Johansson and Flanagan(2009)}]{johansson2009coding}
Johansson R, Flanagan J (2009) {Coding and use of tactile signals from the
  fingertips in object manipulation tasks}. Nature Reviews Neuroscience
  10(5):345--359

\bibitem[{Johnson and Johnson(1987)}]{hertz_contact_mechanics_book}
Johnson K, Johnson K (1987) Normal Contact of Elastic Solids: Hertz Theory,
  Contact Mechanics, chap~4

\bibitem[{Kaneko and Tanie(1994)}]{kaneko1994contact}
Kaneko M, Tanie K (1994) Contact point detection for grasping an unknown object
  using self-posture changeability. IEEE Transactions on Robotics and
  Automation

\bibitem[{Kao et~al(2008)Kao, Lynch, and Burdick}]{contact_springer_handbook}
Kao I, Lynch K, Burdick JW (2008) Contact Modeling and Manipulation, Springer
  Handbook of Robotics, chap~27

\bibitem[{Kavraki and LaValle(2008)}]{planning_springer_handbook}
Kavraki LE, LaValle SM (2008) Chapter 5: Motion Planning, Handbook of Robotics,
  Siciliano, Bruno; Khatib, Oussama (Eds.). Springer

\bibitem[{Khatib(1987)}]{khatib1987unified}
Khatib O (1987) {A unified approach for motion and force control of robot
  manipulators: The operational space formulation}. IEEE Journal of Robotics
  and Automation

\bibitem[{Killpack et~al(2010)Killpack, Deyle, Anderson, and
  Kemp}]{killpack2010visual}
Killpack M, Deyle T, Anderson C, Kemp C (2010) Visual odometry and control for
  an omnidirectional mobile robot with a downward-facing camera. In: IEEE/RSJ
  International Conference on Intelligent Robots and Systems

\bibitem[{Kim and Streit(1995)}]{kim1995configuration}
Kim H, Streit D (1995) Configuration dependent stiffness of the puma 560
  manipulator: analytical and experimental results. Mechanism and machine
  theory

\bibitem[{Kroshko(2011)}]{openopt}
Kroshko DL (2011) Openopt framework. \urlprefix\url{http://openopt.org}

\bibitem[{Kulchenko and Todorov(2011)}]{kulchenko2011first}
Kulchenko P, Todorov E (2011) First-exit model predictive control of fast
  discontinuous dynamics: Application to ball bouncing. In: IEEE International
  Conference on Robotics and Automation (ICRA)

\bibitem[{LaValle and Kuffner(2001)}]{lavalle2001randomized}
LaValle S, Kuffner J (2001) {Randomized kinodynamic planning}. The
  International Journal of Robotics Research 20(5):378

\bibitem[{Lederman and Klatzky(2009)}]{klatzky2009}
Lederman SJ, Klatzky RL (2009) Haptic perception: A tutorial. Attention,
  Perception and Psychophysics (71):1439--1459

\bibitem[{Leeper et~al(2012)Leeper, Hsiao, Ciocarlie, Takayama, and
  Gossow}]{leeper2011strategies}
Leeper A, Hsiao K, Ciocarlie M, Takayama L, Gossow D (2012) Strategies for
  human-in-the-loop robotic grasping. In: ACM/IEEE international conference on
  Human Robot Interaction -- {To Appear}

\bibitem[{Legagne et~al(2011)Legagne, Kheddar, and Yoshida}]{Legagne2011}
Legagne S, Kheddar A, Yoshida E (2011) {Generation of Optimal Dynamic
  Multi-Contact Motions : Application to Humanoid Robots}. IEEE Transactions on
  Robotics -- under review

\bibitem[{Lozano-Perez(1987)}]{lozano1987simple}
Lozano-Perez T (1987) A simple motion-planning algorithm for general robot
  manipulators. IEEE Transactions on Robotics and Automation

\bibitem[{Lumpkin et~al(2010)Lumpkin, Marshall, and
  Nelson}]{cell_bio_of_touch2010}
Lumpkin EA, Marshall KL, Nelson AM (2010) The cell biology of touch. Journal of
  Cell Biology 191(2):237--248, \doi{10.1083/jcb.201006074}

\bibitem[{Maladen et~al(2010)Maladen, Ding, Umbanhowar, Kamor, and
  Goldman}]{maladen2010biophysically}
Maladen R, Ding Y, Umbanhowar P, Kamor A, Goldman D (2010) {Biophysically
  inspired development of a sand-swimming robot}. Robotics: Science and Systems
  (RSS)

\bibitem[{Manchester et~al(2011)Manchester, Mettin, Iida, and
  Tedrake}]{manchester2011stable}
Manchester I, Mettin U, Iida F, Tedrake R (2011) Stable dynamic walking over
  uneven terrain. The International Journal of Robotics Research

\bibitem[{Mason et~al(2011)Mason, Rodriguez, Srinivasa, and
  Vazquez}]{mason2011autonomous}
Mason M, Rodriguez A, Srinivasa S, Vazquez A (2011) Autonomous manipulation
  with a general-purpose simple hand. International Journal of Robotics
  Research

\bibitem[{McKenna et~al(2008)McKenna, Anhalt, Bronson, Brown, Schwerin,
  Shammas, and Choset}]{mckenna2008toroidal}
McKenna J, Anhalt D, Bronson F, Brown H, Schwerin M, Shammas E, Choset H (2008)
  Toroidal skin drive for snake robot locomotion. In: International Conference
  on Robotics and Automation

\bibitem[{Metta et~al(2011)Metta, Natale, Nori, and Sandini}]{mettaforce}
Metta G, Natale L, Nori F, Sandini G (2011) Force control and reaching
  movements on the icub humanoid robot. International Symposium on Robotics
  Research

\bibitem[{Migliore(2009)}]{migliore2009}
Migliore S (2009) {The Role of Passive Joint Stiffness and Active Knee Control
  in Robotic Leg Swinging: Applications to Dynamic Walking}. PhD thesis,
  Georgia Institute of Technology

\bibitem[{Migliore et~al(2005)Migliore, Brown, and
  DeWeerth}]{migliore2005biologically}
Migliore S, Brown E, DeWeerth S (2005) Biologically inspired joint stiffness
  control. In: IEEE International Conference on Robotics and Automation

\bibitem[{Morari and Lee(1999)}]{morari1999model}
Morari M, Lee JH (1999) Model predictive control: past, present and future.
  Computers \& Chemical Engineering 23(4-5):667--682

\bibitem[{Natale and Torres-Jara(2006)}]{natale2006sensitive}
Natale L, Torres-Jara E (2006) A sensitive approach to grasping. In:
  International Workshop on Epigenetic Robotics

\bibitem[{Ott et~al(2007)Ott, Baeuml, Borst, and Hirzinger}]{dlr_door}
Ott C, Baeuml B, Borst C, Hirzinger G (2007) Autonomous opening of a door with
  a mobile manipulator: A case study. IFAC Symposium on Intelligent Autonomous
  Vehicles

\bibitem[{Park and Khatib(2008)}]{Park2008}
Park J, Khatib O (2008) {Robot multiple contact control}. Robotica 26(5)

\bibitem[{Pastor et~al(2011)Pastor, Righetti, Kalakrishnan, and
  Schaal}]{pastor2011online}
Pastor P, Righetti L, Kalakrishnan M, Schaal S (2011) Online movement
  adaptation based on previous sensor experiences. In: IROS

\bibitem[{Patil et~al(2011)Patil, van~den Berg, and
  Alterovitz}]{patil2011motion}
Patil S, van~den Berg J, Alterovitz R (2011) Motion planning under uncertainty
  in highly deformable environments. In: Robotics: Science and Systems (RSS)

\bibitem[{Petrovskaya et~al(2007)Petrovskaya, Park, and
  Khatib}]{petrovskaya2007}
Petrovskaya A, Park J, Khatib O (2007) {Probabilistic Estimation of Whole Body
  Contacts for Multi-Contact Robot Control}. IEEE International Conference on
  Robotics and Automation

\bibitem[{Platt~Jr et~al(2003)Platt~Jr, Fagg, and Grupen}]{platt2003extending}
Platt~Jr R, Fagg A, Grupen R (2003) Extending fingertip grasping to whole body
  grasping. In: IEEE International Conference on Robotics and Automation

\bibitem[{Pratt(2002)}]{low_impedance_walking_robots_pratt}
Pratt G (2002) {Low impedance walking robots 1}. Integrative and Comparative
  Biology 42(1):174--181

\bibitem[{Pratt and Williamson(1995)}]{pratt1995series}
Pratt G, Williamson M (1995) {Series elastic actuators}. In: IROS

\bibitem[{Quigley et~al(2009)Quigley, Gerkey, Conley, Faust, Foote, Leibs,
  Eric~Berger, and Ng}]{quigley09}
Quigley M, Gerkey B, Conley K, Faust J, Foote T, Leibs J, Eric~Berger RW, Ng A
  (2009) {ROS: An Open-Source Robot Operating System}. In: {ICRA Open-Source
  Software workshop}

\bibitem[{Raibert and Craig(1981)}]{raibert1981hybrid}
Raibert M, Craig J (1981) {Hybrid position/force control of manipulators}.
  Journal of Dynamic Systems, Measurement, and Control 102(127):126--133

\bibitem[{Raibert et~al(2008)Raibert, Blankespoor, Nelson, Playter
  et~al}]{raibert2008bigdog}
Raibert M, Blankespoor K, Nelson G, Playter R, et~al (2008) Bigdog, the
  rough--terrain quadruped robot. In: Proceedings of the 17th World Congress

\bibitem[{Rodriguez et~al(2006)Rodriguez, Lien, and
  Amato}]{rodriguez2006planning}
Rodriguez S, Lien J, Amato N (2006) Planning motion in completely deformable
  environments. In: IEEE International Conference on Robotics and Automation

\bibitem[{Romano et~al(2011)Romano, Hsiao, Niemeyer, Chitta, and
  Kuchenbecker}]{romano2011human}
Romano J, Hsiao K, Niemeyer G, Chitta S, Kuchenbecker K (2011) Human-inspired
  robotic grasp control with tactile sensing. IEEE Transactions on Robotics

\bibitem[{Salisbury(1984)}]{salisbury1984interpretation}
Salisbury J J (1984) Interpretation of contact geometries from force
  measurements. In: ICRA, vol~1

\bibitem[{Saranli et~al(2001)Saranli, Buehler, and
  Koditschek}]{saranli2001rhex}
Saranli U, Buehler M, Koditschek D (2001) Rhex: A simple and highly mobile
  hexapod robot. The International Journal of Robotics Research

\bibitem[{Saxena et~al(2008)Saxena, Driemeyer, and Ng}]{saxena2008rgn}
Saxena A, Driemeyer J, Ng A (2008) {Robotic Grasping of Novel Objects using
  Vision}. The International Journal of Robotics Research 27(2):157

\bibitem[{Sentis and Khatib(2005)}]{sentis2005swb}
Sentis L, Khatib O (2005) {Synthesis of whole-body behaviors through
  hierarchical control of behavioral primitives}. International Journal of
  Humanoid Robotics

\bibitem[{Sentis et~al(2010)Sentis, Park, and Khatib}]{Sentis2010}
Sentis L, Park J, Khatib O (2010) {Compliant Control of Multicontact and
  Center-of-Mass Behaviors in Humanoid Robots}. IEEE Transactions on Robotics
  26(3)

\bibitem[{Shadmehr(1993)}]{shadmehr1993control}
Shadmehr R (1993) Control of equilibrium position and stiffness through
  postural modules. Journal of motor behavior

\bibitem[{Simon(1996)}]{simon1996sciences}
Simon H (1996) The sciences of the artificial. the MIT Press

\bibitem[{Sims~Jr et~al(1988)Sims~Jr, Cavanagh, and Ulbrecht}]{foot_injury}
Sims~Jr D, Cavanagh P, Ulbrecht J (1988) Risk factors in the diabetic foot.
  Physical Therapy

\bibitem[{Smith et~al(2011)}]{physics_ode}
Smith R, et~al (2011) Open dynamics engine. \urlprefix\url{http://www.ode.org}

\bibitem[{Srinivasa et~al(2009)Srinivasa, Ferguson, Berenson, Collet, Diankov,
  Gallagher, Hollinger, Kuffner, and VandeWeghe}]{srinivasa2009herb}
Srinivasa S, Ferguson C D~Helfrich, Berenson D, Collet A, Diankov R, Gallagher
  G, Hollinger G, Kuffner J, VandeWeghe M (2009) {Herb: A Home Exploring
  Robotic Butler}. Autonomous Robots

\bibitem[{Stilman et~al(2007)Stilman, Schamburek, Kuffner, and
  Asfour}]{stilman2007manipulation}
Stilman M, Schamburek J, Kuffner J, Asfour T (2007) {Manipulation planning
  among movable obstacles}. In: IEEE Int. Conf. on Robotics and Automation

\bibitem[{Stulp et~al(2009)Stulp, Kresse, Maldonado, Ruiz, Fedrizzi, and
  Beetz}]{stulp2009compact}
Stulp F, Kresse I, Maldonado A, Ruiz F, Fedrizzi A, Beetz M (2009) Compact
  models of human reaching motions for robotic control in everyday manipulation
  tasks. In: IEEE International Conference on Development and Learning

\bibitem[{Ulmen et~al(2012)Ulmen, Edsinger, and Cutkosky}]{ulmenskin}
Ulmen J, Edsinger A, Cutkosky M (2012) A highly sensitive, manufacturable,
  low-cost tactile sensor for responsive robots. In: In Submission, IEEE
  International Conference on Robotics and Automation.

\bibitem[{webpage(2011{\natexlab{a}})}]{racoon_eggs}
webpage (2011{\natexlab{a}})
  \urlprefix\url{http://www.nativeamerica.com/research.html}

\bibitem[{webpage(2011{\natexlab{b}})}]{long_macaque}
webpage (2011{\natexlab{b}})
  \urlprefix\url{http://wildshores.blogspot.com/2009/06/wild-monkeys-at-breakfast-in-admiralty.html}

\bibitem[{webpage(2011{\natexlab{c}})}]{noodling}
webpage (2011{\natexlab{c}})
  \urlprefix\url{http://www.ethantw.com/noodling.html,
  http://en.wikipedia.org/wiki/Noodling}

\bibitem[{Wieber(2006)}]{wieber2006trajectory}
Wieber P (2006) Trajectory free linear model predictive control for stable
  walking in the presence of strong perturbations. In: IEEE-RAS International
  Conference on Humanoid Robots

\bibitem[{Williamson(1996)}]{williamson1996ppi}
Williamson M (1996) {Postural primitives: Interactive behavior for a humanoid
  robot arm}. In: Proceedings of the Fourth International Conference on
  Simulation of Adaptive Behavior

\bibitem[{Williamson(1999)}]{williamson1999rac}
Williamson M (1999) {Robot arm control exploiting natural dynamics}. PhD
  thesis, Massachusetts Institute of Technology

\end{thebibliography}

\end{document}